\documentclass[journal]{IEEEtran}

\usepackage{amsmath}
\usepackage{graphicx}
\usepackage{amsfonts}
\usepackage{moreverb,url}
\usepackage{color,soul}
\usepackage[dvipsnames]{xcolor}
\usepackage[caption=false]{subfig}
\usepackage{siunitx}
\DeclareSIUnit{\impulseunit}{N.s}
\newenvironment{remark}[1][Remark]{\begin{trivlist}
\item[\hskip \labelsep {\bfseries #1}]}{\end{trivlist}}

\begin{document}

\title{Walking Control Based on Step Timing Adaptation}

\author{Majid~Khadiv$^{\star}$,
        Alexander~Herzog$^{\nabla}$,
        S.~Ali.~A.~Moosavian$^{\dagger}$,
        and~Ludovic~Righetti$^{\star,\circ}$
\thanks{$^{\star}$Movement Generation and Control Group, Max-Planck Institute for Intelligent Systems, Germany. e-mail: firstname.lastname@tuebingen.mpg.de}
\thanks{$^{\nabla}$Google X, email: alexanderherzog001@googlemail.com}
\thanks{$^{\dagger}$Department of Mechanical Engineering, K. N. Toosi University of Technology, Tehran, Iran. e-mail: moosavian@kntu.ac.ir}
\thanks{$^{\circ}$Department of Electrical and Computer Engineering and Department of Mechanical and Aerospace Engineering, New York University, USA. e-mail: ludovic.righetti@nyu.edu}}

\markboth{Journal of \LaTeX\ Class Files}
{Khadiv \MakeLowercase{\textit{et al.}}:A Robust Walking Controller for Bipedal Locomotion}


\maketitle

\begin{abstract}
\footnote{Part of the material presented in this paper has been presented at the 2016 IEEE/RAS International Conference on Humanoid Robots (Humanoids), \cite{khadiv2016step}}
Step adjustment can improve the gait robustness of biped robots, however
the adaptation of step timing is often neglected as it gives
rise to non-convex problems when optimized over several footsteps.
In this paper, we argue that it is not necessary to optimize walking over several steps to ensure gait viability and show
that it is sufficient to merely select the next step timing and location.
Using this insight, we propose a novel walking pattern generator
that optimally selects step location and timing at every control cycle.
Our approach is computationally simple compared to standard approaches in the literature, yet guarantees that any viable state will remain 
viable in the future.
We propose a swing foot adaptation strategy and integrate the pattern generator with
an inverse dynamics controller that does not explicitly control the center of mass nor the foot center
of pressure.
This is particularly useful for biped robots with limited
control authority over their foot center of pressure, such as robots with point feet or passive ankles.
Extensive simulations on a humanoid robot with passive ankles demonstrate the capabilities of the approach
in various walking situations, including external pushes and foot slippage, and emphasize the importance
of step timing adaptation to stabilize walking. 

\end{abstract}


\IEEEpeerreviewmaketitle

\section{Introduction}

\subsection{Motivation}
\IEEEPARstart{D}{ue} to the unilateral nature of feet-ground interaction, legged robots can easily fall down. 
The most important aspect of walking control is therefore preventing falls even in face of strong perturbations. 
In this regard, a controller can act on three different aspects of the gait: 
1) where to step, 2) when to step, and 3) how to manipulate ground reaction forces.
State of the art walking pattern generators mostly focus on the third aspect, i.e, the optimal modulation of the center of pressure (CoP), 
sometimes in conjunction with step placement or timing adaption. While such strategies afford
great flexibility in walking pattern generation, they implicitly assume high control authority
on the end-effectors' CoP. Furthermore, since adapting step timing often gives rise to non-convex optimization problems,
 most available walking pattern generators keep footstep timing fixed.

In this paper, we study the problem of optimal footstep location and timing adaptation
without explicit control of the feet CoP or the robot center of mass (CoM).
This enables the relaxation of control constraints on the CoP associated to more traditional receding horizon algorithms,
which is particularly relevant for biped robots with no or limited control authority over
their CoP: robots with passive ankles, small or point feet.

\subsection{Realtime walking pattern generators}

%
To date, the most successful approaches for real-time walking control mostly consider linear models of the CoM dynamics and especially the linear inverted pendulum model (LIPM) \cite{kajita20013d}. Indeed, more complex models often lead to high dimensional, non-convex and computationally complex algorithms,
therefore  limiting their applicability for real-time planning and control \cite{mordatch2012discovery,khadiv2017optimal,carpentier2016A,herzog2016structured}. Recent work for quadruped locomotion also consider more descriptive dynamic models than the LIPM while retaining the capability of running algorithms in real-time. For instance, assuming the legs are massless and the base rotation is small, \cite{di2018dynamic} uses a linear dynamic model between the base states and contact forces for real-time motion planning. \cite{grandia2019feedback} proposes an efficient motion planner and controller based on differential dynamic programming (DDP) that uses an approximation of the robot base dynamics. Although these proposed model predictive control (MPC) approaches are real-time capable, they need another reactive planner level to generate contact sequences. There are also formulations based on more descriptive models with the capability to reason about contact sequence and timing, for example using a relaxed contact model \cite{neunert2018whole} or adding implicitly complementary constraints \cite{winkler2018gait}. However, the resulting optimization problems are not convex and can easily get stuck in undesired local minima. In bipedal locomotion, the LIPM is still often preferred because legs constitute a considerable amount of the robot's total mass and there is little need to vary so much the CoM height.

Leveraging analytical solutions of the LIPM, several approaches generate CoM trajectories consistent with a predefined ZMP trajectory \cite{harada2006analytical,morisawa2006biped,buschmann2007collocation}. 
In these approaches, both the position and velocity of the CoM are restricted which constrains both divergent and convergent parts of the LIPM dynamics. In contrast, \cite{takenaka2009real} constrained only the divergent part of the CoM dynamics to generate a trajectory for the divergent component of motion (DCM) based on predefined footprints (ZMP trajectory). 
Prior to \cite{takenaka2009real}, the divergent part of the LIPM dynamics was also used to explain human
walking characteristics under the name of extrapolated center of mass (XCoM) \cite{hof2008extrapolated}.
This concept is equivalent to the original capture point (CP) idea \cite{pratt2006capture}, i.e. the point on which the robot should step to come to a stop.
In these methods, 
there is no feedback from the current state of the robot to adapt the CoM motion in the presence of disturbances. To circumvent this, \cite{englsberger2015three} proposed a feedback law tracking a DCM trajectory. Although this controller can quickly react to disturbances, perfect DCM tracking assumes unconstrained CoP manipulation.

All of these walking pattern generators can be seen as variants of the same MPC scheme \cite{wieber2016modeling}. One of the pioneering work relating walking pattern generation to optimal control \cite{kajita2003biped} proposed a preview control method to generate CoM trajectories based on predefined ZMP trajectories. Feedback from the current robot state was used to recompute and adapt the motion online. \cite{wieber2006trajectory} improved the performance of this approach in the presence of relatively severe pushes, by constraining the motion of the ZMP inside the support polygon rather than predefining a desired ZMP trajectory. The resulting algorithm computes both ZMP and CoM trajectories respecting feasibility constraints at each control cycle.
In all these approaches, the receding horizon is several foot steps long with fixed step locations and timing. 


\subsection{Step adjustment and timing adaptation}
Explicit manipulation of the CoP to control the DCM or CoM imposes several restrictions on the leg design.
Indeed, robots with point contact feet \cite{hubicki2016atrias} or robots with passive ankles \cite{khadiv2016stepping} have very limited control authority, if any, on the foot CoP.
The modulation of the CoP is also very limited for robots with actuated ankles, due to the rather small foot support area and the limited amount of available ankle torque.
This is in contrast to step adjustment, which allows to select the next step location in a relatively large area compared to the support polygon.
It constitutes therefore a more significant tool for stabilizing biped walking.
Step adaptation algorithms makes walking pattern generators more robust against disturbances.
Successful approaches can either employ heuristics to adapt predefined step sequence based on DCM tracking error \cite{englsberger2015three,khadiv2016stepping}, 
consider step location as a decision variable in an MPC framework  \cite{diedam2008online,herdt2010online,herdt2010walking}, 
use constrained optimization to adapt step location \cite{feng2016robust, kamioka2018simultaneous} or  
modify a preview control approach \cite{urata2012online}. 
However, in all these methods step timing is never adapted as it would render the problem non-convex when considered over several footsteps.

One of the earliest work on step timing adaptation \cite{morisawa2006biped} used an analytical approach to generate CoM/CoP trajectories, and a step period adaptation algorithm to alleviate ZMP fluctuations in case of an immediate change of the step location. \cite{nishiwaki2010strategies} proposed a three stage algorithm to adapt ZMP trajectory, step location and timing. First, preview control is used to adapt the ZMP and compensate for CoM tracking error. If the computed ZMP is outside the support polygon, a second stage extends or shortens the step period. Finally the third stage adapts the next step location.
\cite {feng2015online} showed that using a combination of step location and timing adaptation increases significantly the basin of attraction for bipedal locomotion. \cite{missura2013omnidirectional} proposed an analytical method for computing nominal gait variables for a desired walking velocity, and an algorithm for adapting both step location and timing based on heuristics. \cite{castano2016dynamic} proposed an analytical method for step timing and foot placement adaptation based on the CoM state feedback, with a priority given to the sagittal gait. \cite{griffin2017walking} modified the analytical approach of \cite{englsberger2015three} by adjusting both step location and timing using heuristics to compensate for the DCM tracking error. 
In addition to step timing, single support duration can also be adapted \cite{park2006online,khadiv2017online} to 
deal with soon or late landing of the swing foot using contact detection. 

Step duration was also used as an optimization variables in \cite{aftab2012ankle,kryczka2015online}. However, these approaches result in non-convex optimization problems which are computationally expensive and do not guarantee convergence to a global minimum. \cite{maximo2016mixed} proposed an extension of the gait planning approach in \cite{herdt2010online} to adjust step duration. They related the problem through a mixed-integer quadratic program (MIQP) which has combinatorial complexity. In \cite{bohorquez2017adaptive}, a robust approach is proposed to deal with the nonlinearity introduced by adapting step timing in a receding horizon controller. Furthermore, \cite{caron2017dynamic,caron2017make} used timing adaptation to limit the acceleration of the swing foot, during walking on uneven terrains.


\subsection{Viability and capturability constraints}
Viability theory \cite{aubin1991viability} is an appealing framework to discuss the stability
of walking. The viability kernel includes all the states from which it is possible to avoid falling \cite{wieber2002stability}. Computing this kernel is generally not possible but it has 
been argued \cite{wieber2008viability} that it is sufficient to limit an integral of the states of the system over several previewed steps to guarantee long-term walking stability \cite{wieber2016modeling}.
In general, viability can be guaranteed by setting a terminal condition on the states when considering several steps \cite{takenaka2009real,englsberger2015three},  or by minimizing any derivative of the CoM over a sufficiently large horizon \cite{kajita2003biped,wieber2006trajectory,herdt2010online}. 
As a result, the majority of walking pattern generators today consider several steps of preview to optimize walking. 

Capturability is the ability to come to a stop after a certain number of steps and was extensively analyzed for the LIMP and its extension in \cite{koolen2012capturability}.
Based on an N-step capturability analysis, \cite{de2012foot} proposed a heuristic approach that tries to find the gait parameters that bring the robot back to a desired gait cycle by taking the least number of steps. 
In \cite{koolen2012capturability}, in particular, a bound for the infinite-step capturability 
set is computed, which is almost equivalent to computing the viablity set of the LIPM.
It is a very valuable result, as this can guarantee viability without the need to
consider several walking steps and without over-constraining the system, 
therefore significantly simplifying the walking pattern generator problem.
To the best of our knowledge, this bound has never been used to
synthesize a walking pattern generator.


\subsection{Contributions of the paper}
In this paper, we use results characterizing the viability kernel of the LIPM \cite{koolen2012capturability} 
to argue that it is sufficient to only consider the next step location and timing
to ensure walking stability from any viable state\footnote{Throughout the paper, we  use the term  \textit{walking stability} in the sense of \textit{viability} (i.e. weak invariance) \cite{wieber2002stability}. We also define \textit{robustness} as the ability to recover from external impulsive disturbances.}.
This means that adding a longer preview of steps, under the LIPM assumption, does not
improve the ability to reject disturbances as long as both step timing and location
are adapted for the next footstep. 
To the best of our knowledge, this insight has not been exploited in control synthesis, 
as state of the art receding horizon approaches always include sufficient conditions for viability 
over a preview of several steps despite it being unnecessary and computationally more complex.

Consequently, we design a model predictive controller that optimizes the next
step location and, most importantly, timing at every control cycle and guarantees walking stability 
from any currently viable state.
The algorithm consists in solving a quadratic program which contains at least an order of magnitude less decision variables than standard MPC walking algorithms \cite{herdt2010walking}. To the best
of our knowledge, it is the first algorithm able to adapt both step timing and location
while keeping a convex problem.
As our approach does not rely on CoP modulations, it is applicable to any biped robots,
including robots with passive ankles and point feet.

We also propose a  strategy to adapt online Cartesian swing foot trajectories
to follow adapted step timing and location.
Extensive simulations on a full humanoid robot with passive ankles using hierarchical inverse 
dynamics \cite{herzog2016momentum} demonstrate the capabilities of the approach for robots
with limited control authority over their CoP.
In particular, our experiments emphasize the importance of timing adaptation 
during walking.
Interestingly, the full body controller does not require an explicit control of the CoP nor of
the CoM horizontal motion. As a result, our approach allows the robot to handle a significant amount of disturbances including external pushes and foot slippage.

This paper extends our preliminary work \cite{khadiv2016step} in three main directions.
First, we present a complete viability analysis of the LIPM under footstep constraints and show that we can find the viability kernel by taking into account only the next footstep. This enables the formulation of a MPC walking controller with viability guarantees.
Second, we present several push recovery simulations on a full humanoid robot with passive ankle and show that we can stabilize walking in
the presence of strong pushes by merely adjusting step location and timing without any explicit control of the CoP or CoM.
Finally, we present several slippage recovery simulations demonstrating that our approach is robust to stance foot slippage. To the best of our knowledge, slippage recovery for humanoid walking is seldom studied in the literature.

\section{Problem formulation and viability}\label{Fundamentals}
%

\subsection{Fundamentals}

The LIPM constrains the motion of the CoM on a plane by using a telescopic massless link connecting the CoP to the CoM \cite{kajita20013d}. Its dynamics is written as
\begin{equation}
\label{LIPM}
\ddot{x} = \omega_0^2 (x-u_0)
\end{equation}
where $x \in \mathbb{R}^2$ is the CoM horizontal position (CoM height has a fixed value $z_0$), and $u_0\in\mathbb{R}^2$ is the CoP position on the floor. For point contact feet, $u_0$ is identical to the contact point. $\omega_0$ is the natural frequency of the pendulum ($\omega_0=\sqrt[]{g/z_0}$, where $g$ is the gravity constant).
Using the DCM, $\xi=x+\dot{x}/\omega_0$, the LIPM dynamics can also be written as \cite{englsberger2015three}
\begin{subequations}
\label{DCM_CoM}
\begin{align}
&\dot{x} = \omega_0 (\xi-x) \label{DCM_CoM_1} \\ 
&\dot{\xi} = \omega_0 (\xi-u_0) \label{DCM_CoM_2}
\end{align}
\end{subequations}
Equation (\ref{DCM_CoM}) explicitly reveals the stable and unstable parts of the LIPM dynamics, where the CoM converges to the DCM (\ref{DCM_CoM_1}) and the DCM is pushed away by the CoP (\ref{DCM_CoM_2}). Solving (\ref{DCM_CoM_2}) as an initial value problem gives
\begin{equation}
\label{initial_value}
\xi(t) = (\xi_0-u_0) e^{\omega_0 t}+u_0
\end{equation}
and the DCM at the end of a step of duration $T$ is 
\begin{equation}
\label{initial@T}
\xi_T = (\xi_0-u_0) e^{\omega_0 T}+u_0
\end{equation}

This representation is very convenient, as it is sufficient to constrain the DCM motion without
considering the stable part to ensure stable walking, for example by constraining the DCM location 
at the end of a specified time $\xi(T)$ as in \cite{englsberger2015three}. 
In this case, a terminal condition (captured state) is set at the end of a predefined number of steps, and the desired DCM at the end of the current step is recursively computed. 

This is not the only way to keep the DCM from diverging. 
In fact, a legged robot can instantaneously change its CoP location $u_0$ by taking a step \cite{koolen2012capturability} in order to limit the DCM motion. This is for example the 
approach taken in \cite{pratt2006capture}.
In \cite{koolen2012capturability}, it was shown that the $\infty$-step capturability region 
is only a function of the maximum step length and minimum step time. Therefore, any viable state
remains viable as long as the next step timing and location are decided such that the distance between the DCM and next step location does not exceed a certain bound. It is the approach we exploit in this paper. 

\begin{figure}
\centering
\setlength{\belowcaptionskip}{0mm} 
\includegraphics[clip,trim=7.3cm 19.5cm 6cm 4cm,width=8.5cm]{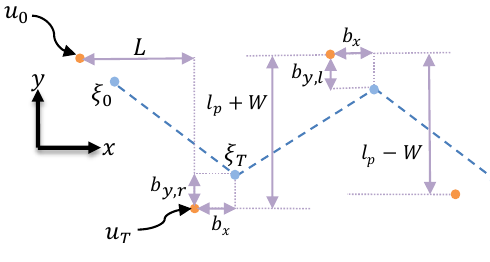}
\caption{Schematic view of walking with footprints,  DCM, and DCM offset.}
\vspace{-0.5em}
\label{DCM}
\end{figure}

\subsection{The DCM offset}
We now introduce the DCM offset, which is a convenient change of variable to synthesize controllers that
enforce a desired CoM average velocity
\begin{equation}
\label{DCM_offset}
b=\xi_T-u_T
\end{equation}
where $u_T$ is the next step location and $\xi_T$ the DCM at the end of the step (Fig. \ref{DCM}). 
The LIPM solution can be written in terms of the next footprint location, the step duration and
the DCM offset by solving \eqref{DCM_CoM_2} as a final value problem
\begin{equation}
\label{final_value}
u_T = (\xi_{cur}-u_0) e^{\omega_0 (T-t)}+u_0-b  \quad ,  \quad  0 \leq t \leq T
\end{equation}
in which $\xi_{cur}$ is the current DCM of the robot.
%
Assume a desired CoM average velocity given by a desired step length $L$, width $W$
and duration $T$, then the resulting desired DCM offset in the sagittal and lateral directions
is simply\footnote{cf. Appendix A for the derivation details}
\begin{subequations}
\label{nominal_offset}
\begin{align}
&b_{x}=\frac{L}{e^{\omega_0 T}-1}\\ 
&b_{y}=(-1)^n\frac{l_p}{1+e^{\omega_0 T}}- \frac{W}{1-e^{\omega_0 T}}
\end{align}
\end{subequations}
where $l_p$ is the default step width and $n=1$ when the right foot is stance, and $n=2$ when the left foot is stance. $W$ is the deviation of the step width with respect to the pelvis width. As this value shows how much the robot moves laterally, we will call it the \textit{step width} (Fig. \ref{DCM}).

\subsection{Viability bound on the DCM offset}
We now express the viability region of the LIPM \cite{aubin1991viability,wieber2002stability,wieber2008viability}
in terms of the DCM offset.
Computing the viability kernel is generally intractable but fortunately, it is possible to characterize these bounds for the LIPM as it was first shown in \cite{koolen2012capturability}, where it was characterized in terms of the $\infty$-step capturability region (i. e. the set of DCM/ICP states from which the system can come to a captured state by taking an infinite (or less) number of steps). 
The  $\infty$-step capturability bound $d_{\infty}$ is written as \cite{koolen2012capturability}
\begin{equation}
d_{\infty} = L_{max} \frac{e^{-\omega_0 T_{min}}}{1- e^{-\omega_0 T_{min}}} \label{eq:inf_step_capt}
\end{equation}
with maximum step length $L_{max}$ and minimum step duration $T_{min}$.
\eqref{eq:inf_step_capt} enables the analysis of the viability of the current state of the system
but is not necessarily convenient to relate viability to the step location and duration. 
%
We now write the viability bounds in terms of the DCM offset.
We limit our analysis to the sagittal plane dynamics for forward walking as the analysis of backward walking 
is similar and the lateral direction analysis is given in Appendix B.  
The maximum DCM offset $b_{x,max}$ is related to the maximum step length and minimum step duration as
\begin{equation}
\label{max_DCM}
b_{x,max}=\frac{L_{max}}{e^{\omega_0 T_{min}}-1} 
\end{equation}
Interestingly, this maximum offset separates viable and non-viable states:
I) if the DCM offset is larger than $b_{x,max}$, all possible choices of step timing and location will lead to its divergence and II) if the DCM offset is smaller than (or equal to) $b_{x,max}$, there exists at least one combination of step timing and location that keeps the DCM from diverging.
\paragraph*{Case I}  If $b_{x,0} > b_{x,max}$ at the start of a step then
\begin{equation}
\label{DCM_cand}
\xi_{x,0}-u_{x,0}=\frac{L_{max}}{e^{\omega_0 T_{min}}-1}+\epsilon
\end{equation}
where $\epsilon>0$. Using \eqref{initial@T}, \eqref{DCM_offset} and \eqref{DCM_cand}, the DCM offset at the end of the step is
\begin{equation}
\label{DCM_end_viable}
b_{x,T} = ( \frac{L_{max}}{e^{\omega_0 T_{min}}-1}+\epsilon ) e^{\omega_0 T}-(u_{x,T}-u_{x,0})
\end{equation}
Substituting $u_{x,T}-u_{x,0}=L_{max}$ and $T=T_{min}$, the minimum realizable offset $b_{x,T}$ is therefore
\begin{equation}
\label{DCM_min}
b_{x,T}=\frac{L_{max}}{e^{\omega_0 T_{min}}-1}+\epsilon \: e^{\omega_0 T_{min}}
\end{equation}
Comparing \eqref{DCM_cand} and \eqref{DCM_min}, we see that the minimum realizable DCM offset at the end of the step increases by $e^{\omega_0 T_{min}}$. 
A sequence of steps will therefore result in a diverging geometric series with common ratio $e^{\omega_0 T_{min}}$ and all possible choices of step location and timing lead to divergence (i.e. a fall).
\paragraph*{Case II} If $b_{x,0} \leq b_{x,max}$ at the start of a step then
\begin{equation}
\label{DCM_less}
\xi_{x,0}-u_{x,0}\leq \frac{L_{max}}{e^{\omega_0 T_{min}}-1}
\end{equation}
Using \eqref{initial@T}, \eqref{DCM_offset} and \eqref{DCM_less}, we find
\begin{equation}
\label{DCM_end_nonviable}
b_{x,T} \leq ( \frac{L_{max}}{e^{\omega_0 T_{min}}-1} ) e^{\omega_0 T}-(u_{x,T}-u_{x,0})
\end{equation}
Selecting the next step position and timing as $u_{x,T}-u_{x,0}=L_{max}$ and $T=T_{min}$, we find
\begin{equation}
\label{DCM_lim}
b_{x,T} \leq \frac{L_{max}}{e^{\omega_0 T_{min}}-1}
\end{equation}
which shows that for any state satisfying  \eqref{DCM_less}, there exists at least one choice of step position and timing that keeps 
the next DCM offset bounded by $b_{x,max}$. The state is viable.

\begin{remark}
The DCM offset bound (\ref{max_DCM}) is equal to the $\infty$-step capturability bound (\ref{eq:inf_step_capt}) derived in \cite{koolen2012capturability}. However,
it affords a different interpretation: the viability kernel defined by the DCM offset bound is the closure of the $\infty$-step capturability set. It
contains states that never lead to a fall but that are not $\infty$-step capturable because they cannot lead to a stop even in an infinite number of steps.
\end{remark}

\section{Stepping controller}\label{Stepping controller}
\begin{figure}
\centering
\includegraphics[clip,trim=4.3cm 1.2cm 11.2cm 5.8cm,width=.35\textwidth]{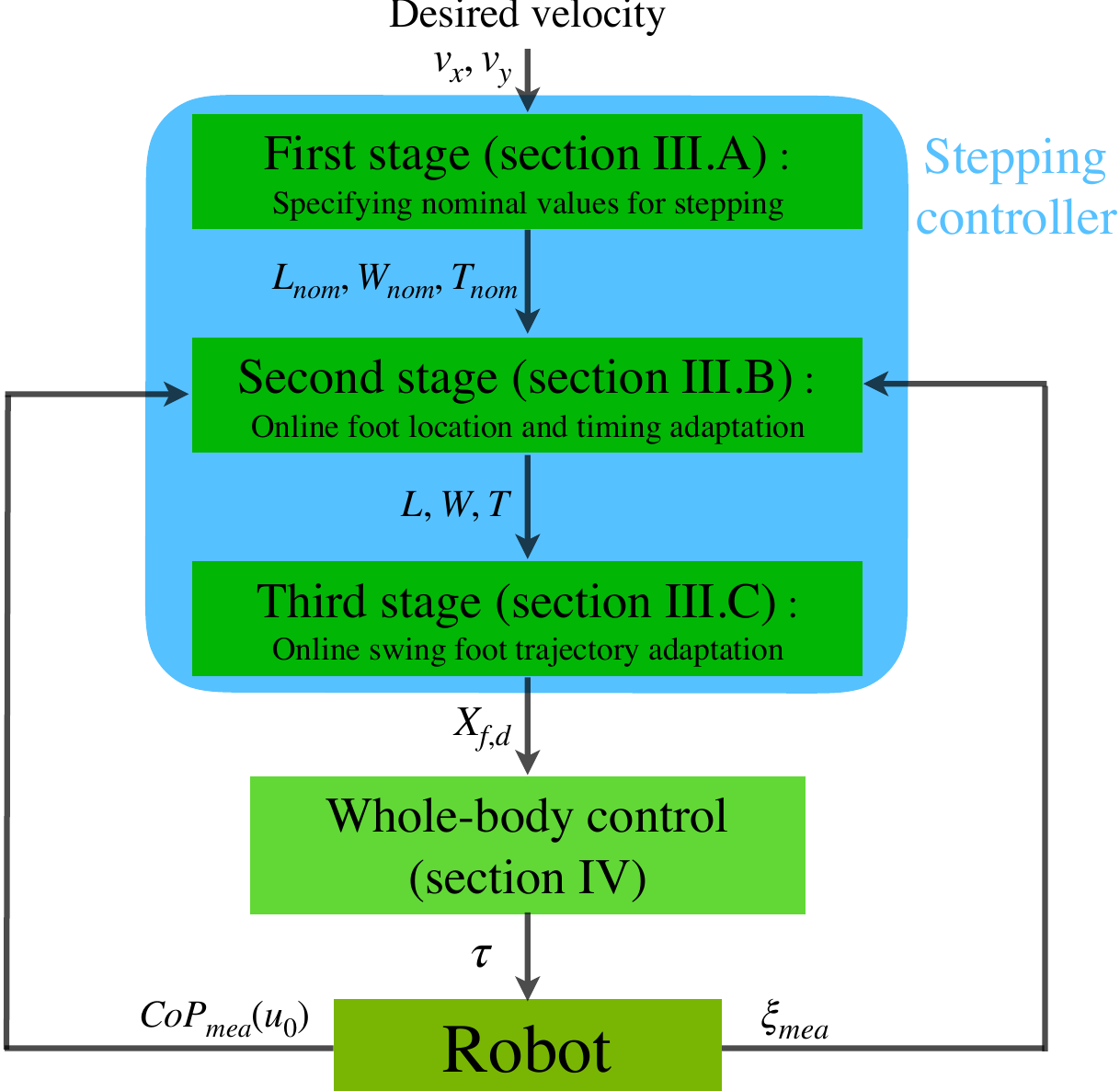}
\caption{Block diagram of the walking algorithm.}
\vspace{-1em}
\label{block_diagram}
\end{figure}

The stepping controller (Fig. \ref{block_diagram}) has three main stages: 
1) we compute nominal step location and duration, and DCM offset for a desired walking velocity,
2) using these nominal values, we compute optimal desired step duration and location at each control cycle and 
3) we adapt the swing foot trajectories used in a whole body controller to generate walking (Sec. \ref{Whole Body Control}).
\subsection{First stage: nominal values for stepping}\label{sec:nominal_values}
We find desired set of step length, width and duration that ensure a desired average walking velocity while 
satisfying the robot and environment constraints. The problem can be formulated in terms of nominal step length $L_{nom}$, width $W_{nom}$, and duration $T_{nom}$ as
\begin{align}
\label{ineq&eq}
&v_x = \frac{L_{nom}}{T_{nom}} \quad, \quad v_y = \frac{W_{nom}}{T_{nom}}\nonumber \\
&L_{min} \leq L_{nom} \leq L_{max}\\ 
&W_{min} \leq W_{nom} \leq W_{max} \nonumber \\
&T_{min} \leq T_{nom} \leq T_{max}\nonumber
\end{align}
where $v_x$ and $v_y$ are the desired average walking velocities in sagittal and lateral directions. 
Step location bounds, $L_{min}$, $L_{max}$, $W_{min}$ and $W_{max}$, 
are set according to the robot (limited step length and width) and environment (limited area for stepping) limitations. Minimum step timing limits 
swing foot acceleration and maximum step timing limits slow stepping.

We compute these nominal values to keep them as far as possible from their boundaries
\begin{align}
\label{nominal_values}
T_{nom}=\frac{B_l+B_u}{2}&\\ 
L_{nom}=v_x(\frac{B_l+B_u}{2})\quad,\quad
W_{nom}&=v_y(\frac{B_l+B_u}{2})\nonumber
\end{align}
where we chose 
\begin{align}
& B_l = \max \{\frac{L_{min}}{|v_x|}, \frac{W_{min}}{|v_y|}, T_{min} \} \quad,\quad v_x,v_y \neq 0\nonumber\\
& B_u = \min \{\frac{L_{max}}{|v_x|}, \frac{W_{max}}{|v_y|}, T_{max} \} \quad,\quad v_x,v_y \neq 0 \nonumber
\end{align}
and we remove the term including $v_x$ or $v_y $ from these equations ,when they are equal to zero. The resulting desired DCM offsets are then computed using \eqref{nominal_offset}.

\subsection{Second stage: online foot location and timing adaptation}
This stage adapts foot step timing and location based on the DCM measurement at each control cycle (typically 1 KHz on a torque controlled robot) to
enforce viability constraints.
Introducing the following change of variable for step timing 
\begin{equation}
\label{variable_changing}
\tau=e^{\omega_0T}\quad \textrm{or}\quad T=\frac{1}{\omega_0} \log(\tau)
\end{equation}
Equation \eqref {final_value} becomes linear in the decision variables $\tau$, $b$, $u_T$
\begin{equation}
\label{linear_equality}
u_T - (\xi_{mea}-u_0) e^{-\omega_0 t} \tau+b=u_0 \quad , \quad 0 \leq t \leq T 
\end{equation}
where $t$ is the time elapsed since the beginning of the actual swing phase. 
The step adaptation problem is then the solution to the quadratic program 
\begin{align}
\label{QP}
\underset{u_{T}, \tau, b}{\text{argmin}} \quad  & \alpha_1 \Vert u_{T}-u_0- \begin{bmatrix} L_{nom}\\W_{nom}\end{bmatrix}  \Vert^2  + \alpha_2 |\tau-\tau_{nom}|^2 \nonumber\\
& + \; \alpha_3  \Vert b-\begin{bmatrix} b_{x,nom}\\b_{y,nom}\end{bmatrix} \Vert^2
 \nonumber\\ \nonumber \\
\text{s.t.} \qquad &\begin{bmatrix} L_{min}\\W_{min} \end{bmatrix}\leq u_T-u_0 \leq  \begin{bmatrix} L_{max}\\W_{max} \end{bmatrix}\nonumber\\
&e^{\omega_0 T_{min}}\leq \tau \leq  e^{\omega_0 T_{max}}\\
&u_T+b=(\xi_{mea}-u_0) e^{-\omega_0 t}\tau+u_0       \nonumber\\
 &\begin{bmatrix} b_{x,min}\\b_{y,max,out} \end{bmatrix}\leq b \leq \begin{bmatrix} b_{x,max}\\b_{y,max,in} \end{bmatrix} \nonumber
\end{align}
where $b_{x,nom}$ and $b_{y,nom}$ are the nominal DCM offsets in sagital and lateral direction computed in Sec. \ref{sec:nominal_values}.
The last inequality in (\ref{QP}) is the viability condition implemented as a hard constraint (cf. Appendix B for $b_{y,max,in}$ and $b_{y,max,out}$ definitions). 
It guarantees that the system remains viable provided that the current state is viable. 
%
In practice, we implement this constraint as a soft constraint with a very high weight with respect to the other cost terms to guarantee that the program will always return a solution even if viability cannot be maintained.
This ensures that measurement noise or differences between the LIPM and the real robot do not lead to a controller failure and enables
to subsequently activate a fall management strategy is necessary.
Lexicographic optimization \cite{escande2014hierarchical} could also be employed in this case but this makes little numerical difference.

\subsection{Third stage: Swing foot trajectory adaptation}\label{Swing foot trajectory adaptation}
We adapt foot trajectories to follow changing step location and timing.
Foot trajectories horizontal to the ground are simply represented with fifth order polynomials so trajectories are continuous in acceleration, which is important for
inverse dynamics control.
%
%
%
%
In the vertical direction, the swing foot height increases until the middle of the step and then decreases to land on the ground. 
Simply using two fifth order polynomials for each part of this trajectory would lead to two important problems: 
1) a change in step timing close to mid-time can cause a jump from the first spline to the second one and 2) a change of step timing in the second part generates unavoidable fluctuations in the vertical direction which may cause ground-foot collisions.
We use instead a 9th order polynomial for the whole step where we keep the swing foot height at the mid-time of the step as close as possible to the desired step height and enforce a stricly positive foot height that is lower than a maximum height. This results in the following QP
\begin{align}
\label{QP_poly}        
\underset{c_i}{\text{argmin}} &\quad \Vert z_{f,d}(T/2)-z_{des} \Vert^2 \\
\quad \text{s.t.} \quad & 0 \leq z(t) \leq z_{max}\nonumber\\
&z_{f,d}(0)=0 \quad z_{f,d}(t_{k-1})=z_{k-1} \quad z_{f,d}(T)=0\nonumber\\ 
&\dot z_{f,d}(0)=0 \quad \dot z_{f,d}(t_{k-1})=\dot z_{k-1} \quad \dot z_{f,d}(T)=0\nonumber\\ 
&\ddot z_{f,d}(0)=0 \quad \ddot z_{f,d}(t_{k-1})=\ddot z_{k-1} \quad \ddot z_{f,d}(T)=0\nonumber    
\end{align}
where $z_{f,d}$ is the vertical component of the swing foot, $k-1$ is the previous sample time and  $T$ is the adapted step timing. 
The coefficients $c_i$ of the polynomial are computed at each control cycle and we evaluate the polynomial at the current time to obtain the current desired foot position.
\section{Whole body control}\label{Whole Body Control}
%
%
We use hierarchical inverse dynamics to control the desired swing foot motion, while enforcing robot constraints, non-slipping contacts and constant CoM height.
We specify a hierarchy of desired task space behaviors and constraints 
expressed as linear equalities and inequalities and the controller computes the resulting optimal joint accelerations, contact forces and actuation torques.
%
We describe in the following our control tasks and constraints,  all details of the controller can be found in \cite{herzog2016momentum}.
\subsection{Foot trajectory tracking}
The swing foot tracking task ensures tracking of the desired motions computed in Section \ref{Swing foot trajectory adaptation}. The task is written as
\begin{equation}
\label{swing_control}
J_{sw} \ddot{q}+\dot{J}_{sw} \dot{q}=\ddot{X}_{f,d}+K_d(\dot{X}_{f,d}-\dot{X}_f)+K_p(X_{f,d}-X_f)
\end{equation}
where $\dot{q}$ is the configuration of the robot, $X_f$ and $X_{f,d}$ are the actual and reference swing foot positions, and $J_{sw}$ is the swing foot Jacobian. $K_p$ and $K_d$ are diagonal gain matrices. We do not control swing foot orientation because we will apply approach to a robot with passive ankles without enough DOFs to control the orientation. For the stance foot, the task is to remain on the ground
\begin{equation}
\label{stance_control}
J_{st} \ddot{q}+\dot{J}_{st} \dot{q}=0
\end{equation}
where $J_{st}$ is the stance foot Jacobian. This task keeps the stance foot in a stationary contact with the ground surface.
\subsection{Center of Mass height tracking}
While it is not necessary to control the horizontal motion of the center of mass, we want to keep a desired CoM height with the following task
\begin{equation}
\label{CoM_control}
J_{CoM} \ddot{q}+\dot{J}_{CoM} \dot{q}=K_p(z_d-z)-K_d \dot{z}
\end{equation}
where $J_{CoM}$ is the CoM height Jacobian, $z$ is the CoM height and $z_d$ is set to the CoM height used in the LIPM ($z_d=z_0$).

\subsection{Posture control}
We ensure a straight robot posture with the task
\begin{equation}
\label{joint_control}
\ddot{q}=K_p(q_d-q)-K_d \dot{q}
\end{equation}
where $K_p$ and $K_d$ are non zero diagonal gain matrices for the actuated joints.
\subsection{Force regularization}
During the double support phase, we exploit contact force redundancy to find forces that distribute the contact forces among the end-effectors in contact
by defining the task objective $\lambda=F_{des}$, where $F_{des}$ is the desired contact force. We put this task in the lowest priority (see Table \ref{hierarchy}) such that acts only as a regularizer when relevant.

\subsection{Task hierarchy}
The hierarchy used for all our experiments is shown in Table \ref{hierarchy}.
We give the highest priority to physical consistency and actuation limits. 
The second rank enforces stance foot contact constraints and CoM height. 
The swing foot control, which essentially generates the walking motions is in the third rank.
The last two ranks are used for the posture control and force regularization.

\begin{table}[h]
\centering
\caption{Hierarchy of tasks used in the humanoid simulation.}
\begin{tabular}{lll}
\hline
Rank&Nr. of eq/ineq constraints&Constraint/Task\\
\hline
\texttt 1 & 6 eq & Centroidal momentum dynamics \\
\texttt   & 2{$\times$}4 ineq & Torque limits \\
\texttt 2 & 6 eq & Stance foot constraint \\
\texttt   & 1 eq & CoM height control \\
\texttt 3 & 3 eq & Swing foot control \\
\texttt 4 & 2{$\times$}4 eq & Posture control \\
\texttt 5 & 2{$\times$}6 eq & Force regularization \\
\hline
\end{tabular}
\label{hierarchy}
\end{table}

\section{LIPM simulation}\label{Abstract model simulation}
In this section, we present simulation results using the LIPM simulation, where \eqref{LIPM} is integrated and $u_0$ is reset by $u_T$ at time $T$ computed from \eqref{QP}. In our implementation of \eqref{QP}, we use a large $\alpha_3$ compared to $\alpha_1$ and $\alpha_2$ to favor solutions close to the desired walking speed. 
For all reported experiments, we set $\alpha_1 = 1$, $\alpha_2 = 5$, $\alpha_3 = 1000$.
Since we optimize the time of the current step at each control cycle, we need to prevent instantaneous stepping (and prevent step timing that are shorter than the current step). Therefore, we apply the adapted gait variables if $T \geq t+T_{gap}$, where $t$ is the current time and $T_{gap}$ is a user defined time gap.
Otherwise we use the previous step values. Additionally, towards the end of a step, when the time gap is smaller than the remaining step time, we stop foot location or timing adaptation until the end of the step. We noticed in our experiments that this was sufficient to avoid time adaptation jitter.\\
The main goal in this section is to show the performance of our gait controller independent of the whole body controller.
We study push recovery capabilities and show the importance of step timing adaptation.
We then compare our controller with a standard MPC-based walking controller \cite{herdt2010online}.

\subsection{Simulation results using the LIPM}
We simulate the LIPM with our controller using a set desired velocity. 
Footsteps and swing foot trajectories are computed as described in Section \ref{Stepping controller}. During each step, the stance foot is set as the point of contact of the LIPM. At the end of a step where the point of contact of the LIPM is changed, the foot index $n$ is changed.
We apply pushes on the robot and compare the recovery capabilities of the controller with and without step timing adjustment.
We set the mass to 60 Kg and the CoM height to 80 cm, Table \ref{physical_properties} defines the other variables (the step location limitations are specified with respect to the stance foot) which correspond to the Sarcos humanoid robot used in the next section.
\begin{table}[h]
\centering
\caption{Physical properties of the abstract model.\label{T1}}
\begin{tabular}{llll}
\hline
Value&Description&min&max\\
\hline
\texttt{$L$} & Step length & $\SI{-50}{cm}$ & $\SI{50}{cm}$\\
\texttt{$W_{right}$} & Step width (right) & $\SI{-10}{cm}$ & $\SI{20}{cm}$\\
\texttt{$W_{left}$} & Step width (left) & $\SI{-20}{cm}$ & $\SI{10}{cm}$\\
\texttt{$T$} & Step duration & $\SI{0.2}{s}$ & $\SI{0.6}{s}$\\
\hline
\end{tabular}
\label{physical_properties}
\end{table}

%
We set a forward desired velocity ($v_x=\SI{1}{m s^{-1}}$) and compute the nominal step length and step duration using \eqref{nominal_values}, as well as the nominal DCM offset using \eqref{nominal_offset}. After four steps, the robot is pushed at $t=\SI{1.4}{s}$ to the right direction with a force $F=\SI{325}{N}$, during $\Delta t = \SI{0.1}{s}$. We conduct two simulations to compare the results of fixed and optimized step duration. In one case, we use \eqref{final_value} for the step  adjustment, using the current DCM measurement. In a second case, we solve the optimization problem in \eqref{QP} at each control cycle to generate the desired time and location of the next step.

\begin{figure}
	\centering
	\setlength{\belowcaptionskip}{0mm} 
	\includegraphics[clip,trim=5.5cm 5.8cm 10cm 1.cm,width=0.48\textwidth]{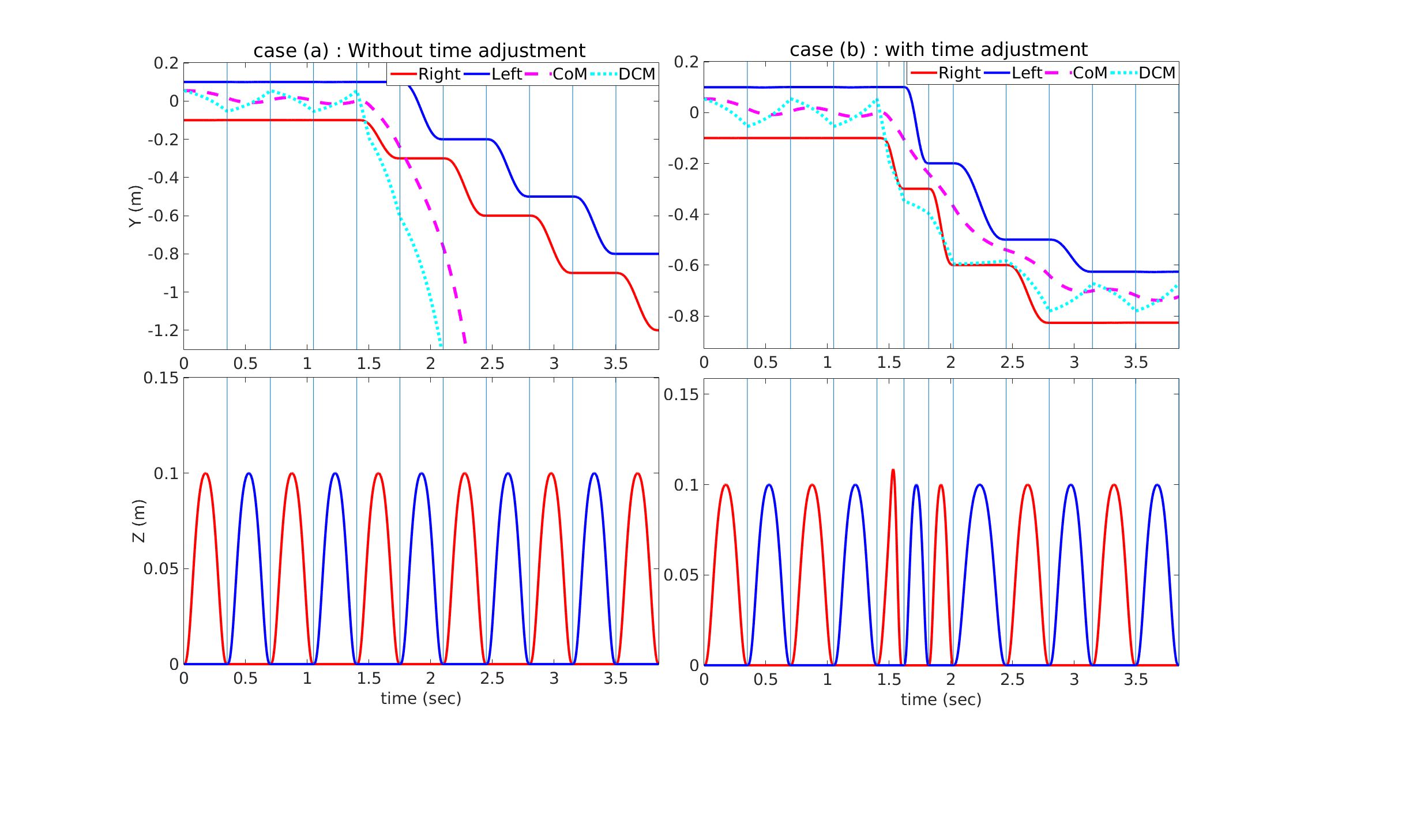}
	\caption{Comparison of trajectories with and without time adjustment. Top: left and right feet, CoM and DCM horizontal positions. Bottom: left and  right feet vertical trajectories. The vertical lines show the step duration. }
	\vspace{-1.5em}
	\label{CoMCoP}
\end{figure}

Figure \ref{CoMCoP} illustrates the resulting trajectories for each case. We observe that without timing adjustment, the robot steps on the borders of the feasible area to recover from the push but because step timing is fixed, the DCM ends up diverging. In the case where timing adjustment is enabled, the algorithm constantly adapts the next footstep and landing time. As a result, the robot steps on the borders of the feasible area very fast to recover from the push. 
We notice that our approach results in smooth swing foot trajectories (without any discontinuity in position or velocity), despite the constant step location and timing adjustment.

\begin{figure}
	\centering
	\setlength{\belowcaptionskip}{0mm} 
	\includegraphics[clip,trim=5cm 1.5cm 6cm 2cm,width=0.48\textwidth]{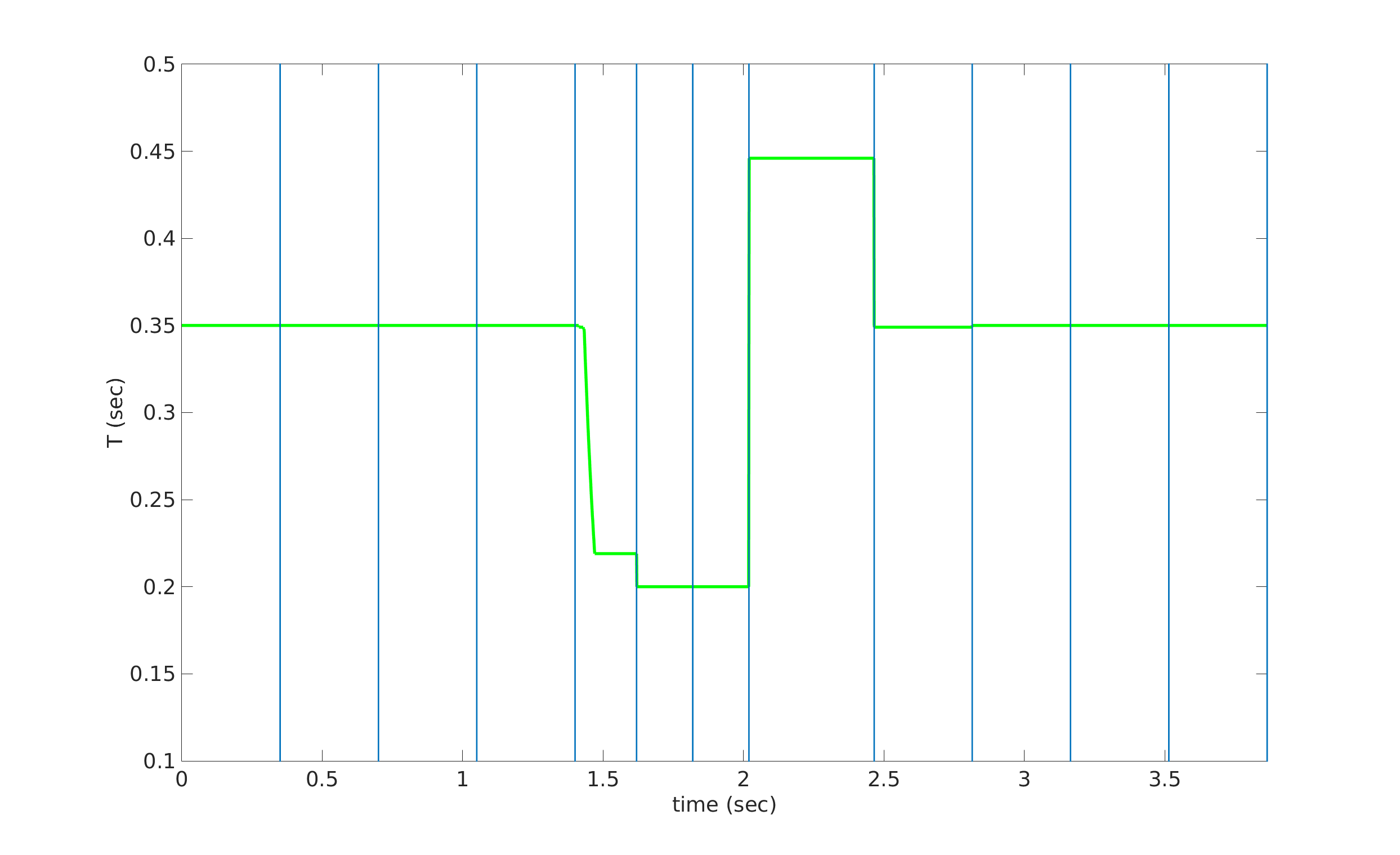}
	\caption{The adapted step time value $T$ versus the simulation time $t$. The vertical lines show the times at which the feet switch.}
	\vspace{-1.5em}
	\label{time_seq}
\end{figure}

Figure \ref{time_seq} shows the adapted step time at each control cycle. The nominal stepping time for this scenario is $T_{nom}=\SI{0.35}{s}$. When the push is exerted at $t=\SI{1.4}{s}$, the optimizer quickly decreases step timing. Interestingly, the optimizer increases step timing after push rejection above the nominal step time ($T=\SI{0.446}{s}$). Indeed,  at $t=\SI{2.019}{s}$ when the push is rejected, the DCM is very close to the stance foot location (Fig. \ref{CoMCoP}, top-right) and the DCM offset far from its nominal value. Hence, the optimizer increases stepping time to bring the robot back to its nominal speed.

\subsection{Comparison with a standard MPC walking controller }
We compare the robustness of our controller to the approach proposed in \cite{herdt2010online}, which 
can be considered, together with variations of this  technique \cite{kajita2003biped,wieber2006trajectory,diedam2008online} as a standard, 
state of the art, walking pattern generator. 
This controller automatically generates step locations and CoM trajectory for a desired walking velocity. Step timing, however, remains fixed. 
In the paper, a horizon of $N=16$ time intervals of length $T=\SI{0.1}{s}$  is used. 
We note here 
that solving the quadratic program for this problem with a 1.6 second horizon is considerably more expensive than our optimization approach.
Indeed, considering CoM jerk and foot location as decision variables in 16 time step horizon yields $4 \times 16 =64$ decision variables at each control cycle without timing adaptation, while our approach only has 6 decision variables.

\begin{figure}
\centering
\setlength{\belowcaptionskip}{0mm} 
\includegraphics[clip,trim=11cm 1.5cm 9.8cm 0.5cm,width=0.48\textwidth]{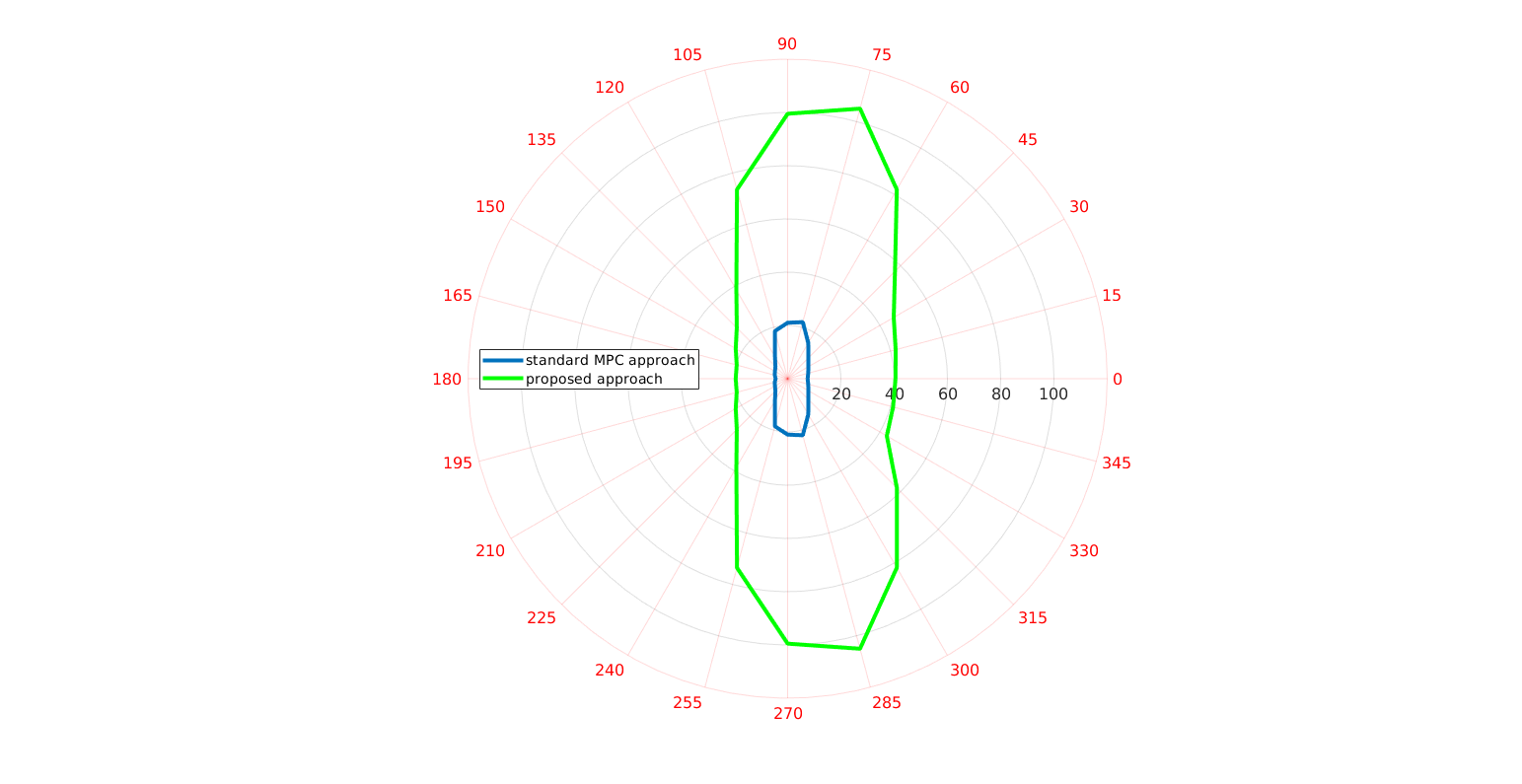}
\caption{Comparison of maximum push impulse (in $\SI{}{\impulseunit}$) before falling. $\theta=\SI{90}{deg}$ corresponds to a forward push, while $\theta=\SI{0}{deg}$ and $\theta=\SI{180}{deg}$ represent pushes to the right and left directions, respectively.}
\vspace{-1.5em}
\label{comparison}
\end{figure}

We used the same parameters for both approaches using a LIPM with point contact and computed the maximum push that each approach can recover from in various directions (Fig. \ref{comparison}).
For each simulation, a force during $ \Delta t =\SI{0.1}{s}$ is applied at the start of a step in which the left foot is stance. We use $T_{nom}=\SI{0.5}{s}$ (computed from the first stage of our algorithm) for both approaches.

Figure \ref{comparison} shows that our walking controller, which only adapts the next step location and time, can recover from much 
more severe pushes compared to the approach with a preview of several steps but without step timing adaptation.
For side pushes, the direction of the push affects the maximum value of the push that the robot is able to withstand. Indeed, for certain pushes (outward direction, left in this case) the feasible stepping area is more limited than the other direction (inward direction) due to leg self-collisions. The maximum push in the direction $\theta =\SI{75}{deg}$ is larger than the other directions because the feasible area for stepping in this direction is larger than the other directions.

We also tested the controller \cite{herdt2010online} with minimum step timing $T_{min}$ as nominal behavior and, as expected, it leads
to the exact same robustness results as our approach. In the case of a very severe push, both algorithms yield stepping on the boundaries of the feasible area. 
This confirms our claim that it is sufficient to solely consider the next foot step timing and location. Moreover, 
in our approach step timing is kept as close as possible to the nominal step time and the shortest step time is only selected when necessary, which should lead
to more desirable behaviors in general. Indeed, 
constantly stepping with minimum step time puts an extra strain on the system. 
Actuators will work at their maximum capacity all the time, augmenting the risk of failure and energy consumption.
Furthermore, increasing the frequency of stepping can make the system more vulnerable to complex terrains with surface unevenness. 
In contrast, our algorithm steps at maximum speed only when absolutely necessary.


\section{Humanoid with passive ankles simulation}\label{Full humanoid simulation}
\subsection{Description of the simulations}
In this section, we use a simulation of a Sarcos humanoid with passive ankles in the SL simulation environment \cite{schaal2009sl}. Contacts are simulated with a penalty method with linear springs and dampers. We used 18 contact points for each foot. The controller has only access to the resultant wrench, simulating a 6-axis force sensor. All experiments are performed on a 2.7 GHz intel i7 processor with  16Gb of RAM. 
The quadratic program (\ref{QP}) is solved using a slightly adapted version of the QuadProg++ software, which implements an active set method described in \cite{goldfarb1983numerically}.
Each leg of the robot has 4 active degrees of freedom with passive ankle joints and prosthetic feet. We simulate the passive ankle joints with stiff springs and dampers. Actuation torques are computed using the controller described in Section \ref{Whole Body Control}.
\begin{figure}
	\centering
	\setlength{\belowcaptionskip}{0mm} 
	\captionsetup[subfigure]{labelformat=empty,position=top}
	\subfloat[(1)]{%
		\includegraphics[width=0.1\textwidth, height=0.13\textheight]{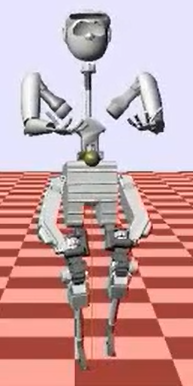}
	}
	\subfloat[(2)]{%
		\includegraphics[width=0.11\textwidth, height=0.13\textheight]{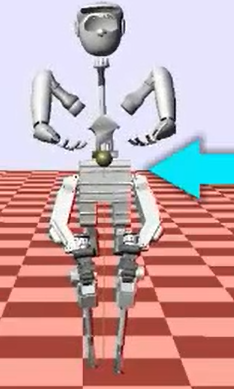}
	}
	\subfloat[(3)]{%
		\includegraphics[width=0.1\textwidth, height=0.13\textheight]{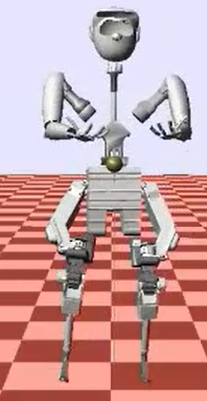}
	}
	\subfloat[(4)]{%
		\includegraphics[width=0.1\textwidth, height=0.13\textheight]{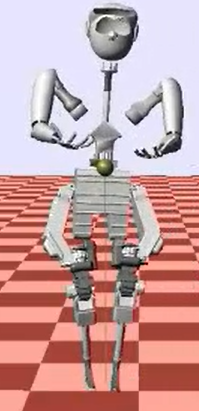}
	}
	
	\subfloat[(5)]{%
		\includegraphics[width=0.1\textwidth, height=0.13\textheight]{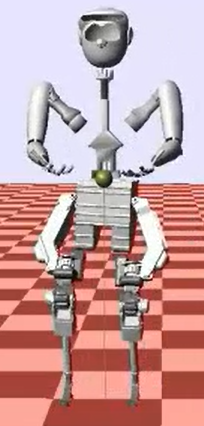}
	}
	\subfloat[(6)]{%
		\includegraphics[width=0.11\textwidth, height=0.13\textheight]{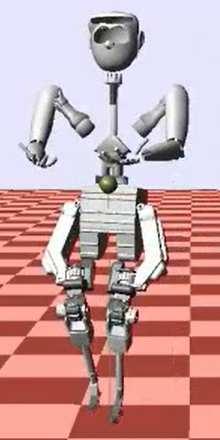}
	}
	\subfloat[(7)]{%
		\includegraphics[width=0.1\textwidth, height=0.13\textheight]{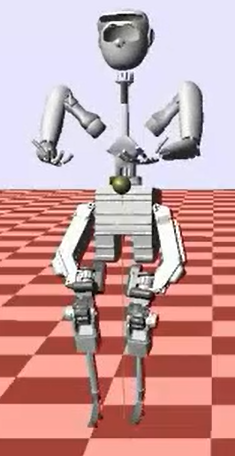}
	}
	\subfloat[(8)]{%
		\includegraphics[width=0.1\textwidth, height=0.13\textheight]{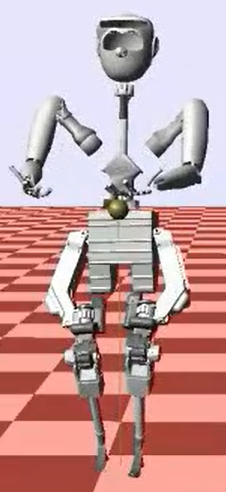}
	}
	\caption{Push recovery example: the robot walks forward at $v_x=\SI{0.2}{ms^{-1}}$, the pelvis is pushed at $t=\SI{3.7}{s}$ by $F=\SI{200}{N}$ during $\Delta t= \SI{0.1}{s}$.}
	\vspace{-1em}
	\label{athena_simulation}
\end{figure}


We conduct different simulations with various external disturbances. Since the robot has finite size feet (i.e. the CoP can move inside the foot), 
we set the current contact point $u_0$ in our controller as the current CoP measurement.
The constraints and physical properties are the same as for the LIPM experiments, except that the minimum step duration is now set to $T_{min}=\SI{0.3}{s}$ to account for acceleration limits of the robot. 
We test two scenarios: push recovery and slippage recovery. In the first scenario, the pelvis is pushed during stepping in different directions. In the second scenario, the stance foot is pushed such that slippage occurs.

\subsection{Push recovery}
We compare our controller with fixed stepping duration
and then with timing adaptation to better understand the influence of step timing adaptation on 
push recovery capabilities.
\begin{figure}
	\centering
	\includegraphics[clip,trim=.8cm .0cm 1.1cm 0.5cm,width=0.48\textwidth]{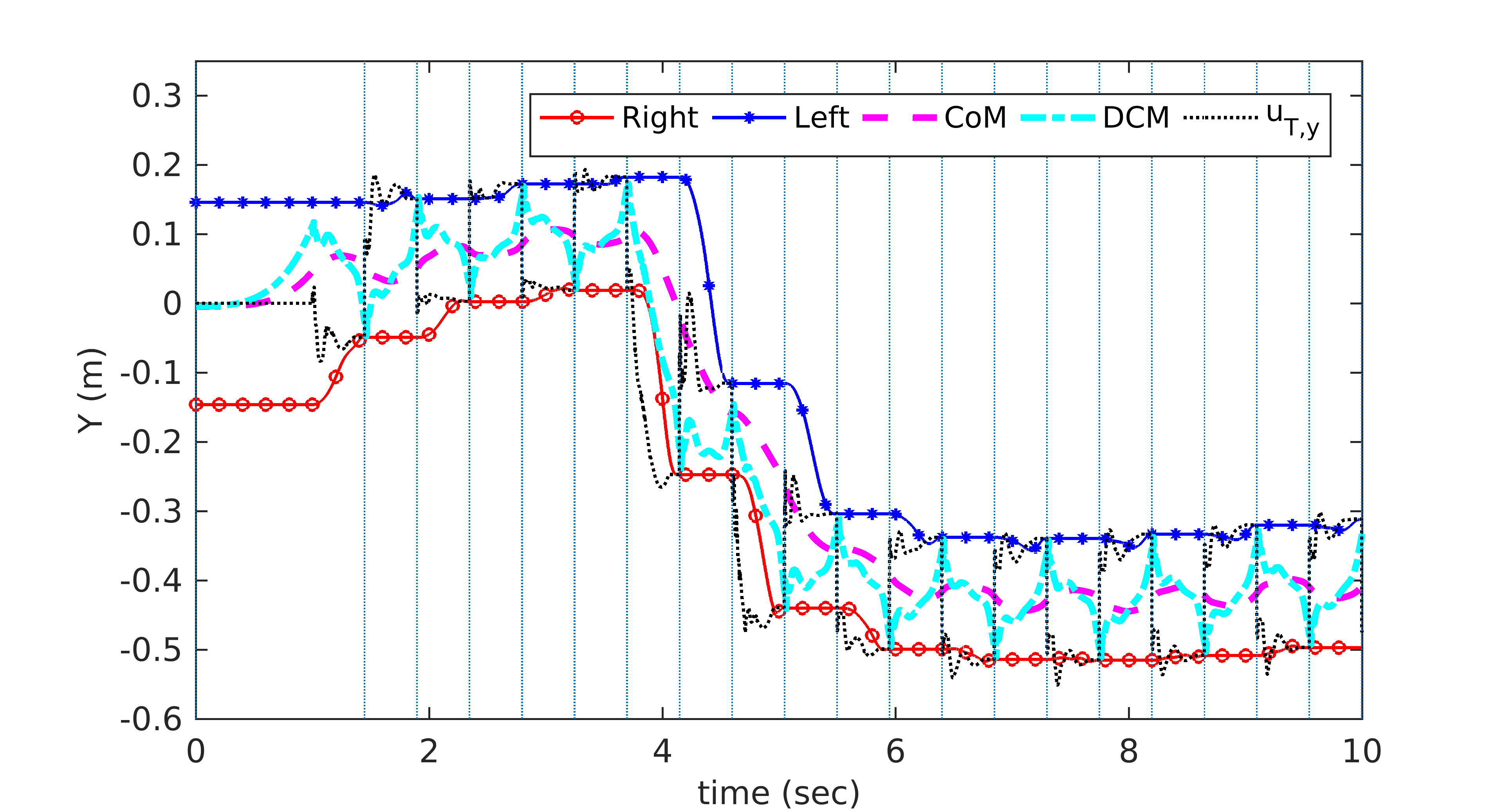}
	\caption{First push recovery experiment: lateral trajectories during forward walking without step timing adjustment. The desired lateral velocity is zero in this simulation, however, when the push occurs ( $F=\SI{200}{N}$, at $t=\SI{3.7}{s}$ during $\Delta t= \SI{0.1}{s}$ which causes an impulse of 20 N.s), the controller sacrifices lateral velocity tracking to recover balance. This push (impulse of $\SI{20}{\impulseunit}$) is the maximum lateral (inward) disturbance that the robot could recover from without timing adjustment.}
	\vspace{-1.em}
	\label{CoMCoP_sim}
\end{figure}
\begin{figure}
	\centering
	\includegraphics[clip,trim=.3cm .1cm 1.1cm .5cm,width=0.48\textwidth]{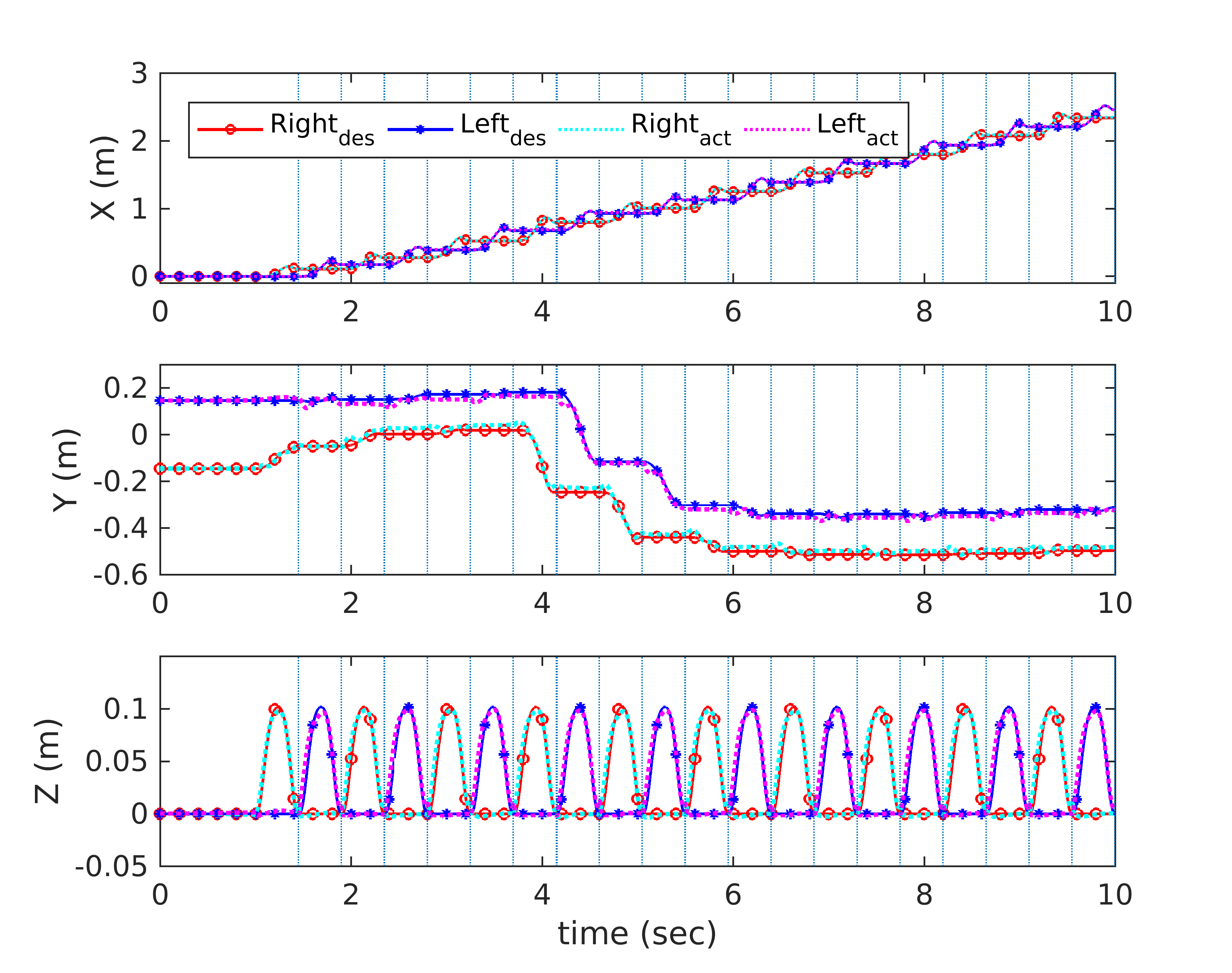}
	\caption{First push recovery experiment: desired and actual feet trajectories during forward walking without step timing adjustment. The low-level controller tracks the smooth feet trajectories generated with the walking controller. }
	\vspace{-1.5em}
	\label{foot}
\end{figure}
Fig. \ref{athena_simulation} shows a typical side-step adaptation behavior. 
To recover from the push, the robot automatically starts stepping to the right direction. Once the push is rejected, the robot resumes its forward walking. 

Fig. \ref{CoMCoP_sim} and Fig. \ref{foot} show push recovery capabilities without time adaptation.
The maximum lateral (inward) disturbance the robot can withstand is $\SI{200}{N}$ for
$\Delta t = \SI{0.1}{s}$. We notice that feet trajectories are adapted to reach the desired landing locations. 
After the push, the controller sacrifices lateral velocity tracking and adjusts foot positions to recover balance. 
The adapted feet trajectories remain smooth resulting in 
smooth control input generated by the whole body controller. 

\begin{figure}
	\centering
	\includegraphics[clip,trim=.8cm .0cm 1.1cm 0.5cm,width=0.48\textwidth]{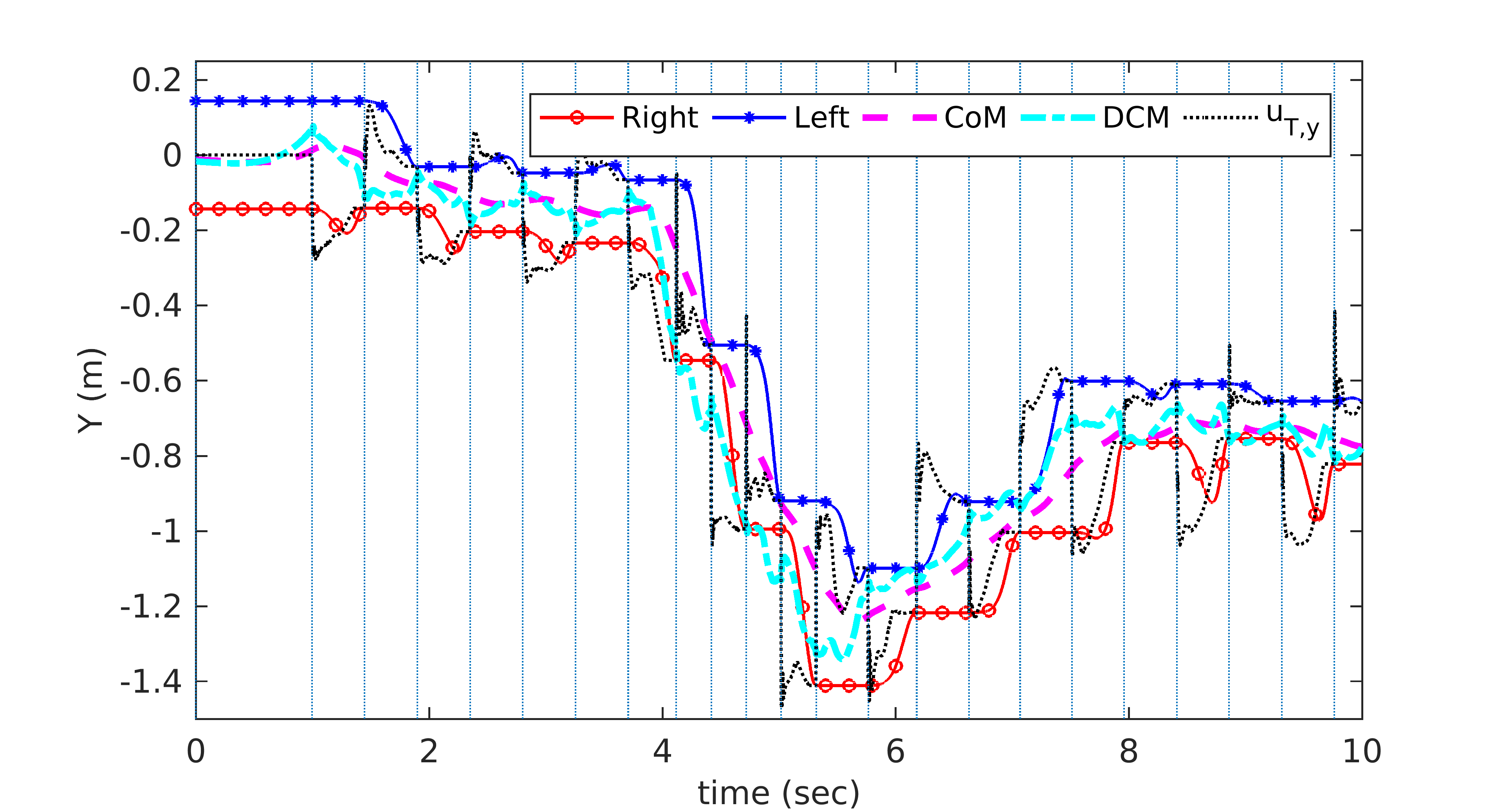}
	\caption{Second push recovery experiment: The lateral trajectories during stepping in place with both step location and timing adjustment. The desired velocity is zero during this simulation. When the push ($F=\SI{390}{N}$, at $t=\SI{3.9}{s}$ during $\Delta t= \SI{0.3}{s}$) is exerted, the controller adapts both step location and timing to recover from this severe lateral (inward) push. This disturbance (impulse of $\SI{117}{\impulseunit}$) is the maximum lateral (inward) push that the robot could recover from with step timing adjustment in our simulations.}
	\vspace{-1.em}
	\label{CoMCoP_sim2}
\end{figure}

\begin{figure}
	\centering
	\includegraphics[clip,trim=.3cm .1cm 1.1cm .5cm,width=0.48\textwidth]{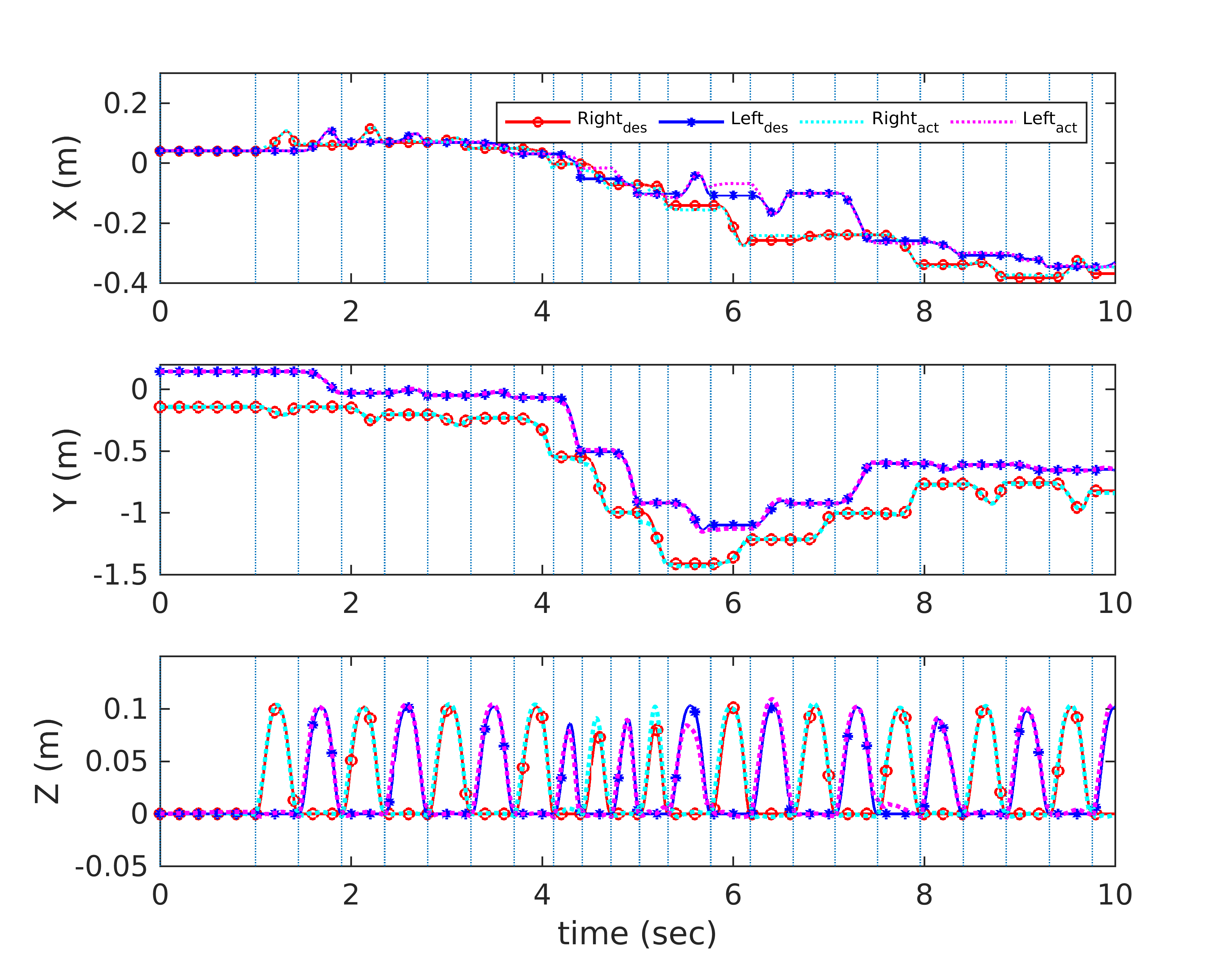}
	\caption{Second push recovery experiment: The desired and actual feet trajectories during stepping in place with both step location and timing adjustment. The low-level controller tracks the smooth feet trajectories generated by our real-time walking controller. We can see that when the step timing is adapted, the step height is adapted to satisfy the constraints specified in (\ref{QP_poly}). }
	\vspace{-1.5em}
	\label{foot2}
\end{figure}

Fig. \ref{CoMCoP_sim2} and Fig. \ref{foot2} show the push recovery results when step timing adaptation is enabled.
In this case, the maximum lateral (inward) disturbance that the robot was able to withstand was a push of $\SI{390}{N}$ for
$\Delta t =  \SI{0.3}{s}$, which is nearly six times the impulse
that the robot could withstand when timing was not adapted. This is consistent with the 
results found for the LIP model (Fig. \ref{comparison} with $\theta$ = \SI{0}{deg}).
This result further illustrates the importance of timing adaptation for stable walking.
Indeed, inspecting \eqref{initial_value} we see that the DCM diverges as an exponential of time, therefore
taking fast steps (decreasing the step timing) enables an exponential improvement in DCM regulation. 
This effect is magnified when the robot takes several steps to recover from a disturbance.

We note a slight difference between these two simulation scenarios. Indeed, the vertical lines in the figures show that for the scenario with timing adaptation the push is exerted in the middle of a step ($t=\SI{3.9}{s}$ in Fig. \ref{CoMCoP_sim2}), while in the scenario without timing adaptation the disturbance is exerted exactly at the start of a step ($t=\SI{3.7}{s}$ in Fig \ref{CoMCoP_sim}). 
Since the DCM diverges exponentially fast, the later the push during a specified step, the more difficult the recovery (e. g. the DCM diverges more when time evolves from 0.2 s to 0.4 s than 0 s to 0.2 s). As a result, although the experimental condition is more difficult in the case of timing adaptation, the robot recovers from a push nearly six times larger than the case without timing adaptation.

Fig. \ref{CoMCoP_sim2} shows that both step location and timing are adjusted. To recover from the push, the robot takes five steps with minimum step time to the left on the boundaries of the feasible area. 
The feet trajectories (Fig. \ref{foot2}) are again adapted very smoothly in the case when step timing is adjusted. Furthermore, the step height is also adapted when step duration is changed, consistently with the constraints specified in \eqref{QP_poly}. Trajectory tracking in the vertical direction degrades compared to the case without timing adjustment (Fig. \ref{foot}), which is due to an increase in the desired swing foot acceleration. However, such trajectory tracking performance in the vertical direction is sufficient for feasible stepping.


\begin{figure}
	\centering
	\includegraphics[clip,trim=.8cm .0cm 1.1cm 0.5cm,width=0.48\textwidth]{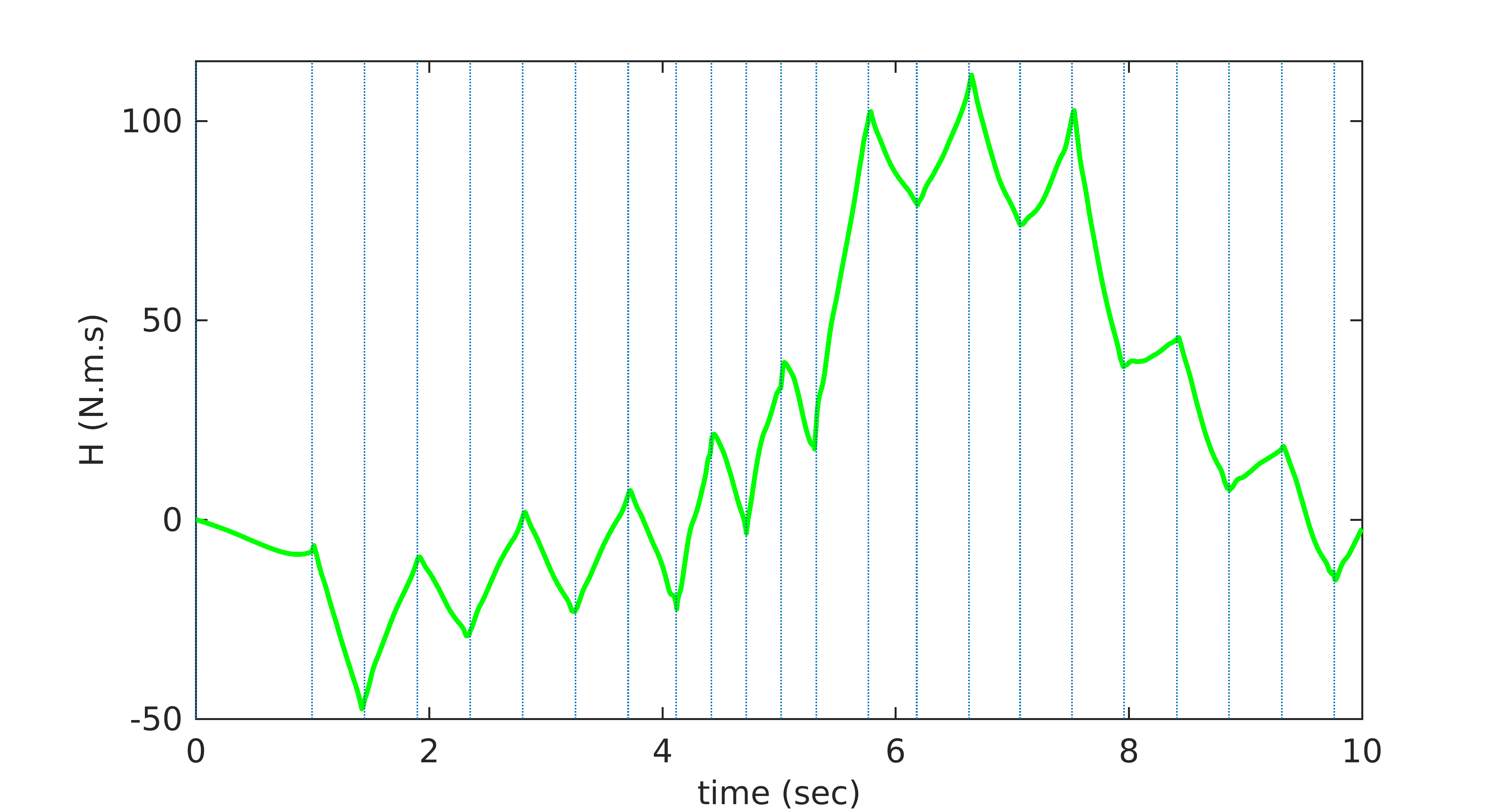}
	\caption{Second push recovery experiment: The angular momentum trajectory around x-axis during stepping in place with both step location and timing adjustment. The desired velocity is zero during this simulation. When the push ($F=\SI{390}{N}$, at $t=\SI{3.9}{s}$ during $\Delta t= \SI{0.3}{s}$) is exerted, the angular momentum is accumulated, until the push is fully rejected (around $t= \SI{6}{s}$). After this, the whole body controller tries to make the angular momentum zero and bring back the joints as close as possible to the configuration where the upper-body is upright. This acts as a kind of disturbance to the system and the robot starts adapting the step in the opposite direction until the angular momentum is around zero and the upper-body is upright.}
	\vspace{-1.em}
	\label{angular_momentum}
\end{figure}

Fig. \ref{CoMCoP_sim2} shows that after rejecting the exerted push by stepping to the right, the robot starts (around $t = \SI{6}{s}$) to step back in left direction. 
This behavior can be explained by looking at the angular momentum (see Fig. \ref{angular_momentum}). The stance foot constraints are in a high rank and the whole body controller sacrifices joint posture and generates angular momentum to keep the stance foot steady on the ground. After exerting the push, angular momentum is accumulated until the push is fully rejected (around $t= \SI{6}{s}$). After this, the whole body controller reduces the angular momentum to zero and brings back the joints toward the desired upright upper-body posture. This acts as a disturbance to the walking controller and the robot starts stepping in the opposite direction until the angular momentum is zero and the upper-body is upright.

%

\begin{figure}
	\centering
	\includegraphics[clip,trim=.8cm .0cm 1.1cm 0.5cm,width=0.48\textwidth]{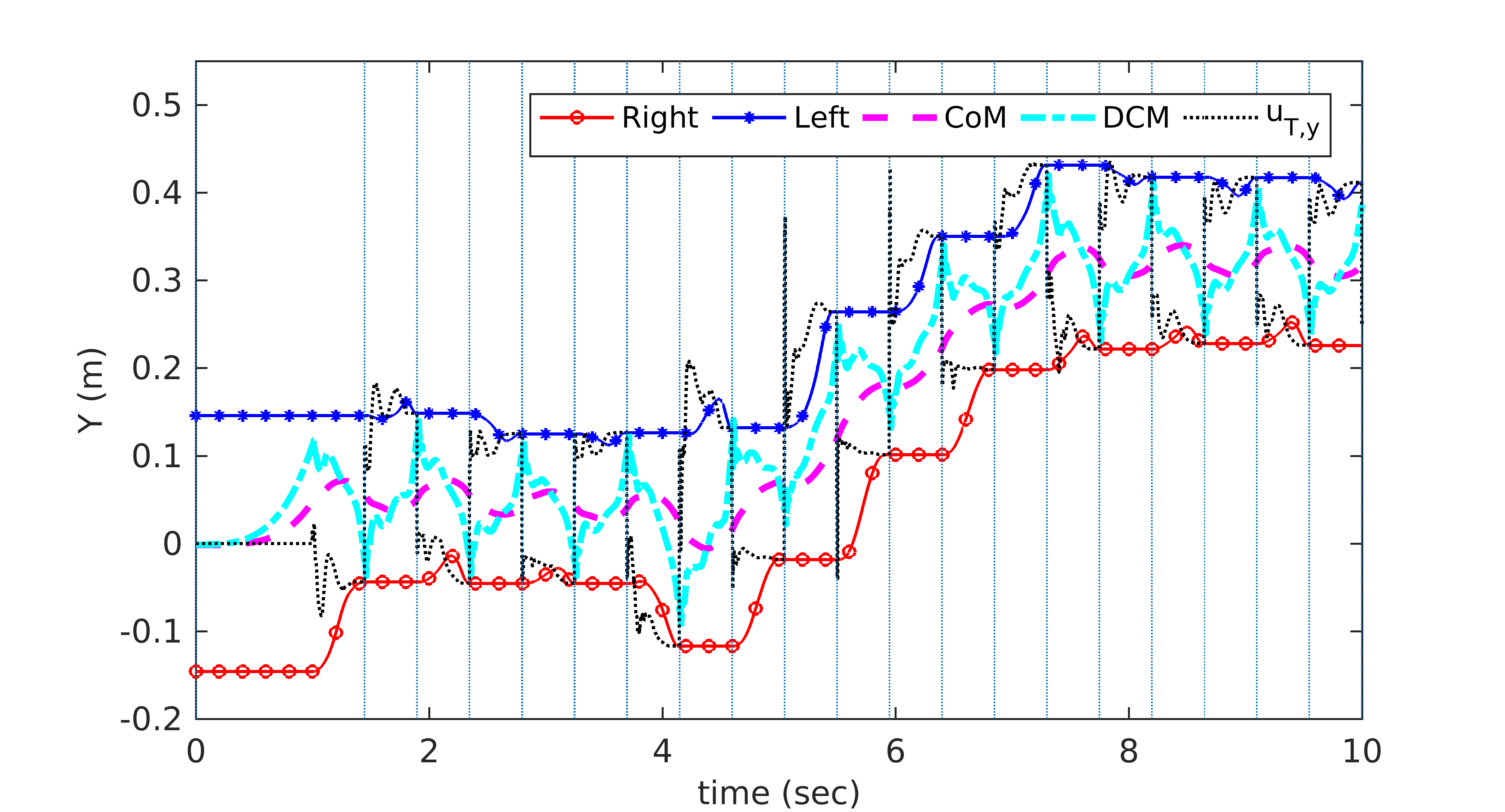}
	\caption{First slippage recovery experiment: The lateral trajectories during a forward walking without step timing adjustment. The desired lateral velocity is zero during this simulation. The stance foot is pushed laterally by $F=\SI{400}{s}$, at $t=\SI{3.9}{s}$ during $\Delta t= \SI{0.2}{s}$ such that slippage occurs. After the push, the foot locations are adjusted to recover the robot from the push.}
	\vspace{-1em}
	\label{CoMCoP_slip}
\end{figure}

\begin{figure}
	\centering
	\includegraphics[clip,trim=.3cm .1cm 1.1cm .5cm,width=0.48\textwidth]{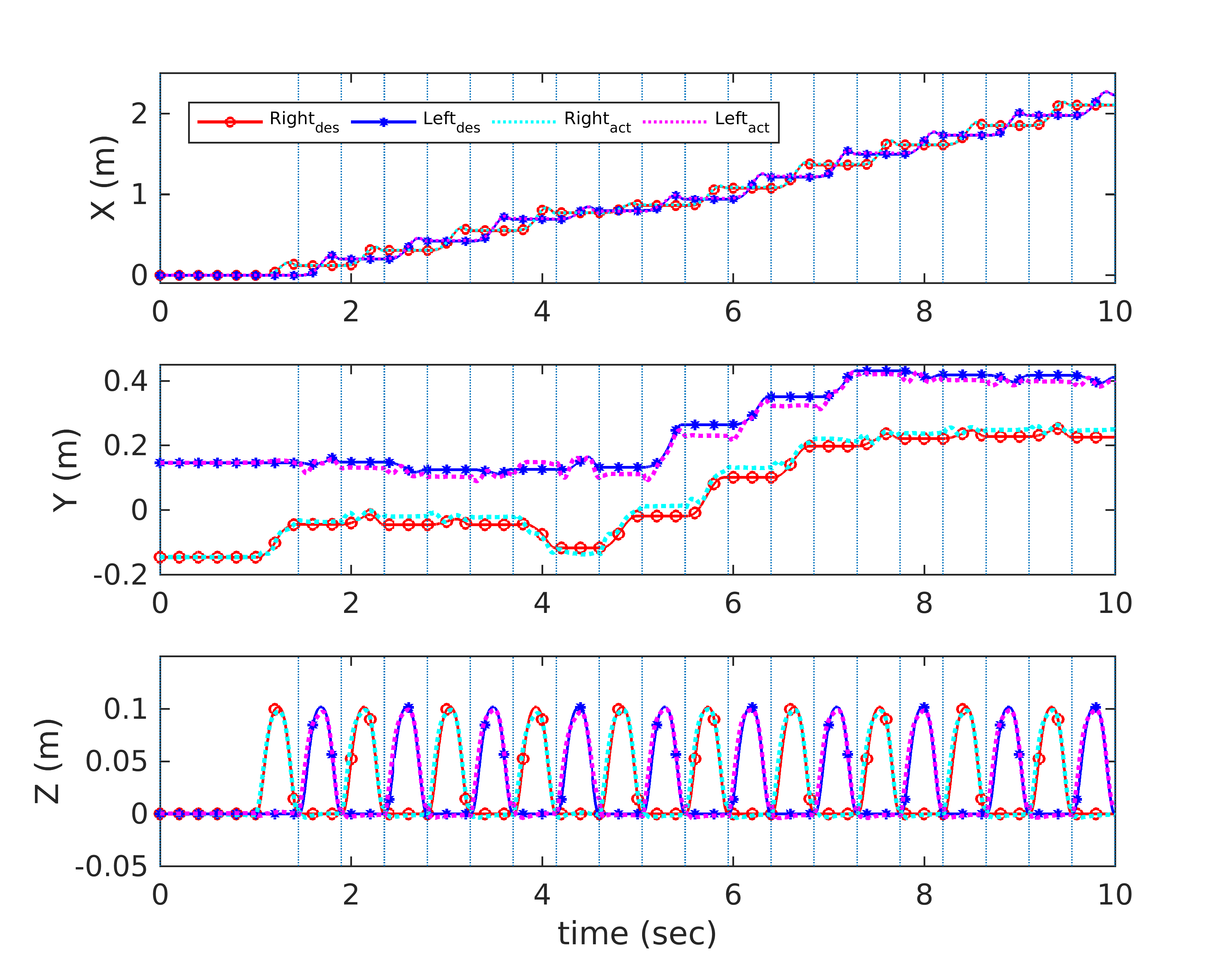}
	\caption{First slippage recovery experiment: The desired and actual feet trajectories during a forward walking without step timing adjustment. When the push is exerted at $t=\SI{3.9}{s}$, the left foot slips to the left. The right foot trajectory is then adapted to recover the robot from this disturbance.}
	\vspace{-1.5em}
	\label{foot_slip}
\end{figure}

\subsection {Slippage recovery}
We now investigate the slippage recovery capabilities of the controller.
Walking controllers based on CoP modulation tend to be very sensitive to foot slippage. Indeed, when the stance foot slips, the assumption of stationary flat foot on the ground is not valid anymore. 
Since our controller does not rely on CoP control it can recover from large slippage by adjusting the swing foot landing location and time.
During walking, we apply strong forces on the stance foot such that it slips. 

\begin{figure}
	\centering
	\includegraphics[clip,trim=.8cm .0cm 1.1cm 0.5cm,width=0.48\textwidth]{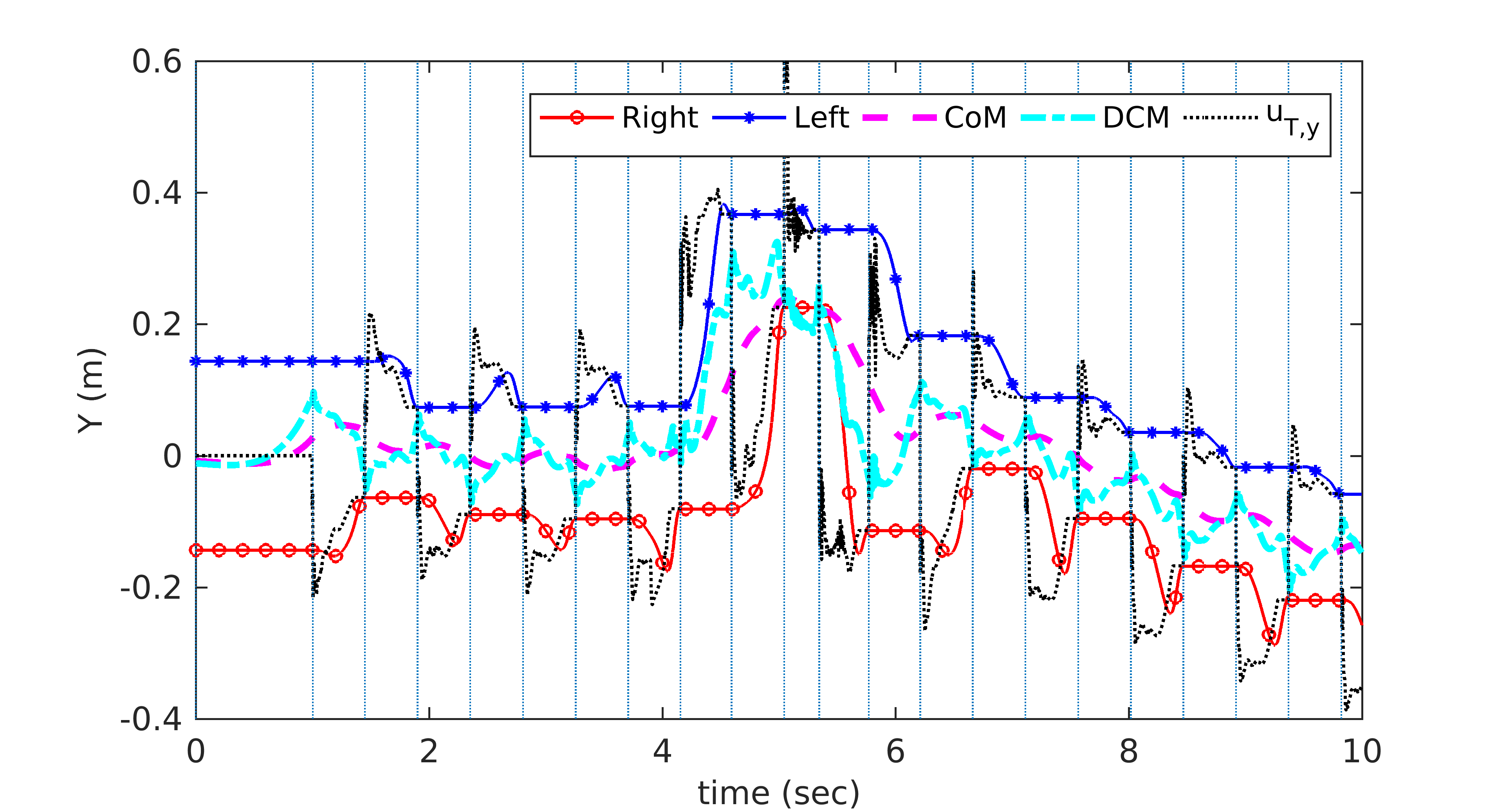}
	\caption{Second slippage recovery experiment: The lateral trajectories during stepping in place with both step location and timing adjustment. The desired velocity is zero during this simulation. The stance foot is pushed laterally by  $F=\SI{930}{N}$, at $t=\SI{3.9}{s}$ during $\Delta t= \SI{0.3}{s}$ such that slippage occurs. After the push, the foot locations are adjusted to recover the robot from the push.High frequency oscillations of the DCM trajectory is due to the huge disturbance on the stance foot. This huge disturbance causes high frequency oscillation of the passive elements in ankles.}
	\vspace{-1.5em}
	\label{CoMCoP_slip2}
\end{figure}

\begin{figure}
	\centering
	\includegraphics[clip,trim=.3cm .1cm 1.1cm .5cm,width=0.48\textwidth]{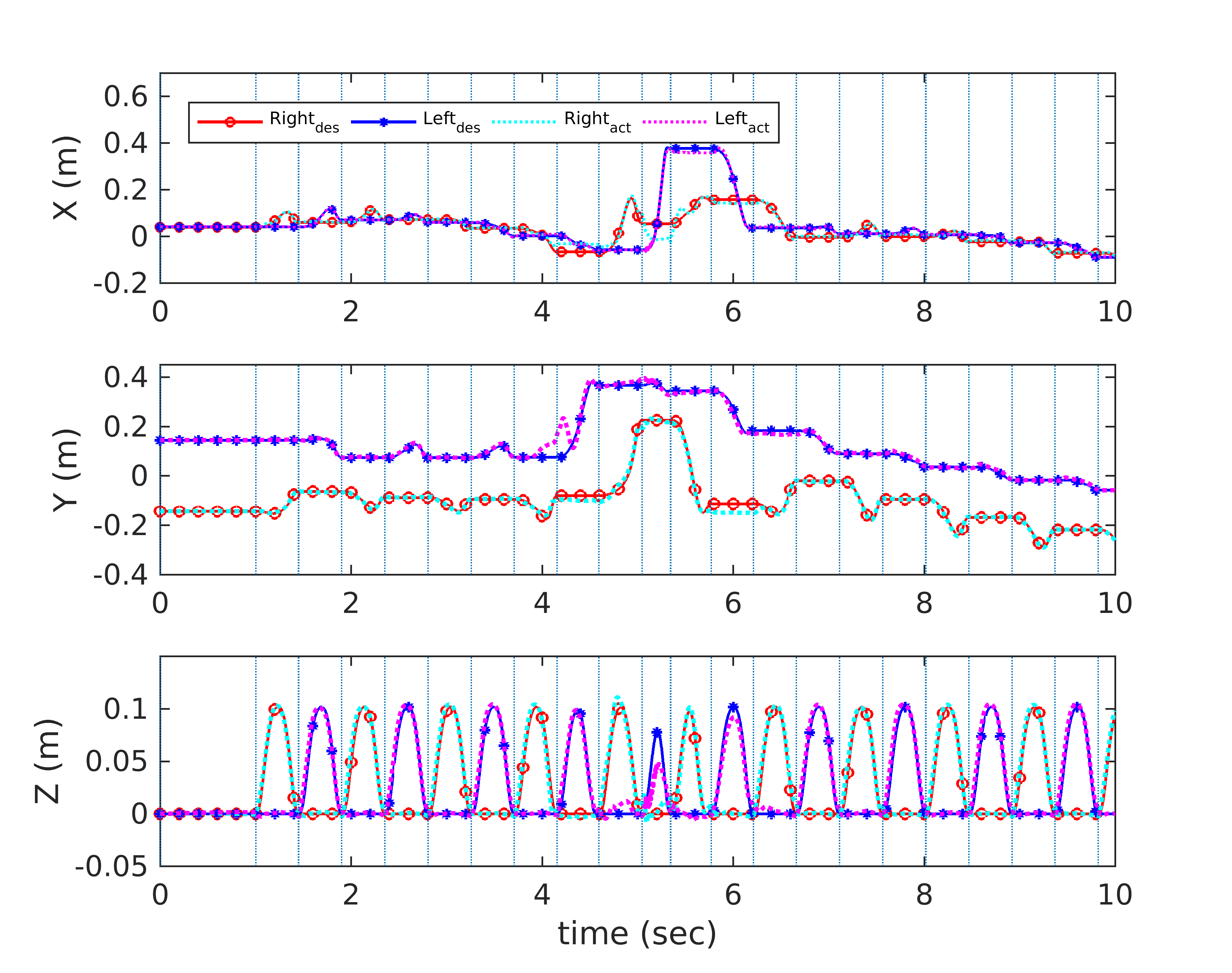}
	\caption{Second slippage recovery experiment:  desired and actual feet trajectories during stepping in place with both step location and timing adjustment. When the push is exerted at $t=\SI{3.9}{s}$, the left foot slips to the left and the next steps locations are adapted to recover from the disturbance.}
	\vspace{-1.5em}
	\label{foot_slip2}
\end{figure} 

We compare performance with and without timing adaptation.
For the case without timing adaptation, Fig. \ref{CoMCoP_slip} shows a typical forward walking simulation. Slippage of the left foot can be observed in the actual foot trajectory in the lateral direction (Fig. \ref{foot_slip}) which causes a faster DCM divergence in the right direction. Hence, the right foot steps further in the right direction to recover from the disturbance.
Figure \ref{CoMCoP_slip2} shows the lateral trajectories when step timing adjustment is enabled. Again, we can see a degradation in the foot trajectory tracking in lateral direction (Fig. \ref{foot_slip2}). The controller exploits a combination of step timing and location adjustment to recover from this strong disturbance. Some high frequency oscillations can be seen in the DCM trajectory and the estimated next footprint (Fig. \ref{CoMCoP_slip2} ), which is due to the large disturbance on the stance foot. This disturbance causes high frequency oscillations of the passive elements in the ankles. Nevertheless, the robot is able to withstand an impulse
on the foot which is more than three times larger than the impulse withstood when step timing  is not adjusted. This again illustrates the importance of timing adjustment during walking.

\section{Discussion}\label{Discussion}
\subsection{Features of the approach}
\begin{itemize}
\item \textit{Generality} Our approach can generate robust gaits without modulating the foot CoP. This suggests that our approach can be used for biped robots with active ankles, passive ankle, or point contact foot. We successfully used the controller to control an underactuated biped robot with passive ankle joints and prosthetic feet. 


\item \textit{Computational efficiency} The size of our optimization problem is drastically smaller than typical MPC approaches using several preview steps \cite{kajita2003biped,wieber2006trajectory,diedam2008online,herdt2010online}, at least an order of magnitude smaller in the number of decision variables.
Viability is ensured by optimizing the DCM offset without the need to integrate the motion forward over several time steps \cite{wieber2016modeling}.

\item \textit{Controller receding horizon length} Our paper provides an answer to an open question for MPC-based controllers: 
\textit{how far ahead in time shall our controller optimize motion for?}\cite{zaytsev2015two}. It is only sufficient to look at the next step to recover walking from any viable state. Interestingly, several reported experiments on human walking seem to support this argument, even in the presence of obstacles \cite{patla1997and,patla2003far}. However, if the terrain is very difficult or constrained (e.g. walking on stepping stones), humans seem to look two steps ahead \cite{patla2003far}. 

\item \textit{Step timing vs step location adjustment}
To highlight the difference between step location and timing adjustment, let's assume that the swing foot can move in all directions without any kinematic constraints and that the only constraint is a limited foot acceleration. Then, the maximum distance the swing foot can travel in a given duration is achieved using a bang-bang controller (maximum acceleration and deceleration) with equal acceleration and deceleration periods if kinematic constraints are not considered. Specifying step period $T$ and maximum acceleration/deceleration value $a_{max}$, the swing foot travels $u_T=a_{max} \; \frac{T^2}{4}$ during this period. 
Assuming $u_0=0$ for the current step in the DCM dynamics, at time $T$ the DCM is $\xi_T = \xi_0 e^{\omega_0 T}$. The DCM diverges as an exponential of time while the next step location can only be adapted with a second order time dependency. 

\item \textit{Robustness against various disturbances}
The measured CoP and estimated DCM are employed to adapt the landing location and time of the swing foot but the controller does not try to control the CoP. This feature enables robust walking under strong pushes and stance foot slippage. This feature might also enable the robot to walk over rough terrains, but this is left as a future work. Furthermore, our experiments showed that timing adjustment significantly increased walking robustness.

\item \textit{Comparison to existing approaches} Step timing adaptation can serve different purposes, i. e. decreasing the ZMP fluctuation in case of immediate change of step location \cite{morisawa2006biped}, walking on uneven terrain \cite{nishiwaki2010strategies, park2006online, khadiv2017online}, disturbance recovery \cite{aftab2012ankle, kryczka2015online, khadiv2016step, griffin2017walking}, respecting acceleration constraint of the swing foot moving on a specified path \cite{caron2017make}, or collision avoidance for walking in crowds \cite{bohorquez2017adaptive}. Despite a variety of goals, the employed approaches are either based on analytical methods and heuristics \cite{park2006online, morisawa2006biped, nishiwaki2010strategies, missura2013omnidirectional, castano2016dynamic, griffin2017walking, khadiv2017online} or optimization  \cite{aftab2012ankle, kryczka2015online, caron2017make, maximo2016mixed, bohorquez2017adaptive}. Optimization provides a principled way of taking into account all constraints while optimizing a desired performance cost. However, non-convexity of the system dynamics with respect to step timing can significantly increase computation time and is prone to getting stuck in undesired local minima. Our approach with at least one order of magnitude less decision variables than existing approaches \cite{aftab2012ankle, kryczka2015online, bohorquez2017adaptive, caron2017make, maximo2016mixed} and a convex formulation circumvents both concerns and provides a reliable tool for realtime applications.
\item \textit{Viability} In \cite{sugihara2017foot}, an approach similar to ours \cite{khadiv2016step} is used for guaranteeing viability of the gait by enforcing a desired offset between the capture point and landing location of the swing foot. However, the approach to realize this offset is based on modulating the CoP of the current step, while the next step location and timing are fixed. 
\cite{santacruz2013reactive} used a combination of ZMP modulation and step location adjustment to keep the capture point inside the support polygon at the end of the current step (1-step capturability). As we showed in this paper, endowing step location adaptation with timing adjustment and using exact viability constraint in the controller structure can ensure that any initially viable state remains viable.
\end{itemize}
\subsection {Limitations}
\begin{itemize}
\item \textit{LIPM assumption}: the viability guarantees of our controller are valid under the assumption of a fixed CoM height and constant angular momentum at the CoM. While these assumptions restrict walking on even ground, since we do not rely on CoP modulation we anticipate that the controller will also be robust to mild rough terrain provided that the low level controller manage the uncertainty in landing. An interesting direction of research would be to use the concept of 3D-DCM and virtual repellent point (VRP) \cite{englsberger2015three} instead to enable our approach to handle 3D maneuvers.
	
\item \textit{Timing constraint} Constraints on step locations are explicit and depend on the kinematic and environment limitations. However, constraints on step timing are a function of maximum acceleration of the swing foot as well as the distance between the current state of the swing foot and the landing location. As a result, considering a  constraint on step timing for the worst case decreases the capabilities of the controller. To circumvent this, we could use more complex, yet linear, models which take into account the swing foot dynamics \cite{takenaka2009real}. Then, we could directly constrain swing foot acceleration instead of step timing. A simpler alternative is to use an empirical linear inequality constraint on the next step location and minimum time as a function of maximum foot speed and  remaining stepping time \cite{herdt2010online}.

\item \textit{Walking on stepping stones or in scattered environments} While we do not need to consider a multi-step horizon to guarantee viability in unconstrained environments, when the feasible foot locations are very limited, such as waking over stepping stones or in the presence of obstacles, we would need to 
consider multiple steps in advance to steer the DCM motion in the direction of safe contact points. 
Our approach can be readily extended to consider this issue, for example one could adapt step timing only in the first step within a desired horizon and optimize for the subsequent step locations with a fixed nominal step timing, therefore retaining a QP.

\item \textit{Real robot experiments} Previous works \cite{feng2015online,pratt2012capturability}
 have shown that our modeling assumptions
can be successfully used on a real robot. While this suggests that our approach would work
as well, real robot experiments, especially for robots with passive ankles or point
feet would be most valuable. At the time when the research was conducted, we did not have
access to such robots.

\end{itemize}

\section{Conclusion}\label{Conclusion}
We proposed a walking controller that adapts both step location and timing in real time to generate robust gaits and we showed that for the LIPM, 
optimizing the next footstep timing and location was sufficient to ensure that viable states remain viable.
Comparison with a standard MPC-based walking controller emphasized the importance of step 
timing adaptation and 
demonstrated that our approach can be significantly more robust (the algorithm could in our simulation tolerate up to about 5 times larger impulsive pushes) than methods that do not adjust timing, even when those controllers optimize over several preview steps. Simulations in push and slippage recovery scenarios on a humanoid with passive ankles showed the robustness of the controller even without control authority in the ankles and the foot CoP.

\section*{Acknowledgment}

This research was supported by the European Unions Horizon 2020 research and innovation program (grant agreement 780684 and European Research Councils grant 637935) and in part by the MPI-ETH center for learning systems. We would like to thank the reviewers for their thoughtful comments on a previous version of the manuscript. 



\bibliography{Master}
\bibliographystyle{IEEEtran}

\appendices
\section*{Appendix A: Derivation of \eqref{nominal_offset}}
 Using \eqref {initial@T} and \eqref {DCM_offset}, the LIPM equation in terms of the DCM offset and the next step location is
\begin{equation}
\label{app_final_value}
u_T = (\xi_0-u_0) e^{\omega_0 T}+u_0-b 
\end{equation}
For a constant walking velocity, the desired DCM offset in the sagittal direction, $b_x$, 
is considered to be constant over several steps, therefore we have 
\begin{equation}
\label{app_DCMx}
u_{T,x} =b_x e^{\omega_0 T}+u_{0,x}-b_x  
\end{equation}
since $L = u_{T,x} - u_{0,x}$ over two consecutive steps, we have
\begin{equation}
\label{app_DCMx_offset}
b_x=\frac{L}{ e^{\omega_0 T}-1} 
\end{equation}
Note that this DCM offset causes a limit cycle for the robot nonlinear (switching) dynamics, in the sense that 
keeping $L$ and $T$ constant over several steps leads to a steady state forward walking velocity.
For a desired sideward walking, the distance between the feet is equal to $l_p+W$ or $l_p-W$ (Fig. \ref{DCM}). Using (\ref{app_final_value}), the equations for the right and left foot DCM offset are
\begin{align}
\label{app_DCMy_dist}
-(l_p-W) &=b_{y,l} e^{\omega_0 T}-b_{y,r} \nonumber \\
l_p+W&=b_{y,r} e^{\omega_0 T}-b_{y,l}
\end{align}
Solving (\ref{app_DCMy_dist}) for $b_{y,r}$ and $b_{y,l}$ yields
\begin{align}
\label{app_DCMy}
&b_{y,r}=\frac{l_p}{1+e^{\omega_0 T}}- \frac{W}{1-e^{\omega_0 T}}\nonumber \\
&b_{y,l}=-\frac{l_p}{1+e^{\omega_0 T}}- \frac{W}{1-e^{\omega_0 T}}
\end{align}

We use the index $n$ to distinguish the right ($n=1$) and left ($n=2)$ stance foot and write the DCM offset in the latteral direction as
\begin{equation}
b_{y}=(-1)^n\frac{l_p}{1+e^{\omega_0 T}}- \frac{W}{1-e^{\omega_0 T}}
\end{equation}

\section*{Appendix B: Viability in the lateral direction}
In the lateral direction, the feasible area for stepping is not symmetric with respect to the stance foot due to self collision. The maximum allowable DCM offsets for outward and inward directions are then different. Here we call outward direction the one where the legs can experience a self collision, and inward for the opposite direction. The maximum allowable DCM offsets in these directions are
\begin{subequations}
\label{app_DCM_all}
\begin{align}
&b_{y,max,out}=\frac{l_p}{1+e^{\omega_0 T_{min}}}+ \frac{W_{max}-W_{min}e^{\omega_0 T_{min}}}{1-e^{2\omega_0 T_{min}}}\label{app_DCM_all_1} \\
&b_{y,max,in}=\frac{l_p}{1+e^{\omega_0 T_{min}}}+ \frac{W_{min}-W_{max}e^{\omega_0 T_{min}}}{1-e^{2\omega_0 T_{min}}} \label{app_DCM_all_2}
\end{align}
\end{subequations}

To show that these values exactly split the state space into the viable and non-viable parts, we again consider the situations where the DCM offsets are less or more than these values for both outward and inward directions. We conduct this analysis for the case where the right foot is stance (the same results directly follow for the other cases).
%
Considering a DCM offset more (less) than the value in (\ref{app_DCM_all_1}) for the outward direction at the start of a step
\begin{align}
\label{app_DCM_max_out}
&\xi_{y,0}-u_{y,0}=\frac{l_p}{1+e^{\omega_0 T_{min}}}+ \frac{W_{max}-W_{min}e^{\omega_0 T_{min}}}{1-e^{2\omega_0 T_{min}}}\pm \epsilon
\end{align}
the minimum feasible DCM offset at the end of this step ($b_{y,T}$) is obtained taking a step with the minimum step width (for avoiding self collision) at the minimum step time using (\ref{app_final_value})
\begin{align}
\label{app_DCM_end_out}
b_{y,T}=&(\frac{l_p}{1+e^{\omega_0 T_{min}}}+ \frac{W_{max}-W_{min}e^{\omega_0 T_{min}}}{1-e^{2\omega_0 T_{min}}}\pm \epsilon)e^{\omega_0 T_{min}}\nonumber\\
&-(l_p+W_{min})
\end{align}
Using this value as the DCM offset at the start of next step, and taking a step with the maximum step width at the minimum step time yield the minimum feasible DCM offset of the next step
\begin{align}
\label{app_DCM_end2_out}
b_{y,2T}&=(\frac{l_p}{1+e^{\omega_0 T_{min}}}+ \frac{W_{max}-W_{min}e^{\omega_0 T_{min}}}{1-e^{2\omega_0 T_{min}}}\pm \epsilon)\nonumber\\
&e^{2\omega_0 T_{min}}-(l_p+W_{min})e^{\omega_0 T_{min}}+(l_p+W_{max})
\end{align}
which simplifies into
\begin{align}
\label{app_simp_end_out}
b_{y,2T}=\frac{l_p}{1+e^{\omega_0 T_{min}}}+ \frac{W_{max}-W_{min}e^{\omega_0 T_{min}}}{1-e^{2\omega_0 T_{min}}} \pm \epsilon e^{2\omega_0 T_{min}}
\end{align}
Comparing (\ref{app_DCM_max_out}) with (\ref{app_simp_end_out}) reveals that starting from a DCM offset more than (\ref{app_DCM_all_1}) (positive $\epsilon$ in (\ref{app_DCM_max_out})) and after taking two steps on the boundaries of the feasible area at the minimum time, the DCM offset increases by a factor $e^{2\omega_0 T_{min}}$ times $\epsilon$. As a result, any DCM offset more than (\ref{app_DCM_all_1}) for causes divergence. However, starting from a DCM offset less than the value in (\ref{app_DCM_all_1}) (negative $\epsilon$ in (\ref{app_DCM_max_out})), there exists one evolution (stepping on the boundaries of the feasible area at the minimum time) that keeps the DCM from diverging. Hence, (\ref{app_DCM_all_1}) for the outward direction exactly splits the state space into viable and non-viable parts. 

%
%
%
%
%
%
%


\begin{IEEEbiography}[{\includegraphics[width=1in,height=1.25in,clip,keepaspectratio]{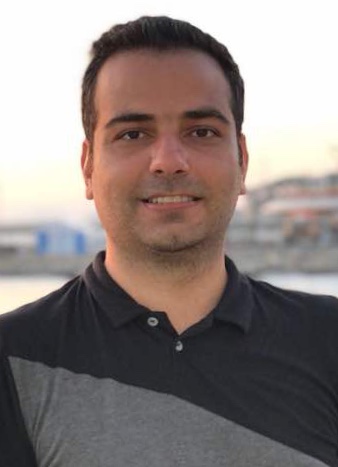}}]{Majid Khadiv}
is a postdoc scholar at the Movement Generation and Control Group at the Max-Planck Institute for Intelligent Systems. He received his BSc degree in Mechanical Engineering from Isfahan University of Technology (IUT), in 2010, and his MSc and PhD degree in Mechanical Engineering from K. N. Toosi University of Technology, Tehran, Iran in 2012 and 2017. Majid joined the Iranian national humanoid project, Surena III, and worked as the head of dynamics and control group from 2012 to 2015. He also spent a one-year visiting scholarship under supervision of Dr. Ludovic Righetti at the Autonomous Motion Laboratory (AMD), Max-Planck Institute for Intelligent Systems. His main research interest is control of legged robots.
\end{IEEEbiography}

\begin{IEEEbiography}[{\includegraphics[width=1in,height=1.25in,clip,keepaspectratio]{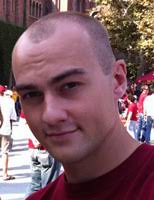}}]{Alexander Herzog}
is a roboticist at X, Inc. (Google). He did his PhD at the Max Planck Institute for Intelligent Systems, T\"ubingen and received his doctorate degree from ETH Z\"urich. Dr Herzog studied Computer-Science at the  Karlsruhe Institute
of Technology, in Germany. He visited the Computational Learning and Motor Control Lab (University of Southern California) regularly from 2011 to 2016 where he collaborated in projects on grasp learning and locomotion for humanoid robots. After receiving his Diploma in 2011, he joined the Autonomous Motion Laboratory at the  Max-Planck Institute for Intelligent Systems in 2012. During his PhD he worked on contact interaction in whole-body control and grasping for humanoids.
\end{IEEEbiography}

\begin{IEEEbiography}[{\includegraphics[width=1in,height=1.25in,clip,keepaspectratio]{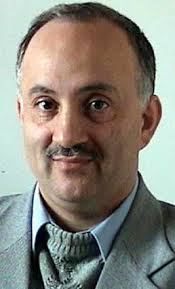}}]{S. Ali. A. Moosavian}
received his B.Sc. degree in 1986 from Sharif University of Technology and the M.Sc. degree in 1990 from  Tarbiat  Modaress  University  (both  in  Tehran),  and  his Ph.D.   degree   in   1996   from   McGill   University   (Montreal, Canada),  all  in  Mechanical  Engineering.  He  is  a  Professor with  the  Mechanical  Engineering  Department  at  K.  N.  Toosi University  of  Technology  (KNTU)  in  Tehran  since  1997. His research interests are in the areas of dynamics modeling and motion/impedance control of terrestrial, legged and space robotic systems. He has published more than 200 articles in peer-reviewed journals and conference proceedings. He is a Member of IEEE, and one of the Founders of the ARAS Research Group, and the Manager of Center of Excellence in Robotics and Control at KNTU.
\end{IEEEbiography}

\begin{IEEEbiography}[{\includegraphics[width=1in,height=1.25in,clip,keepaspectratio]{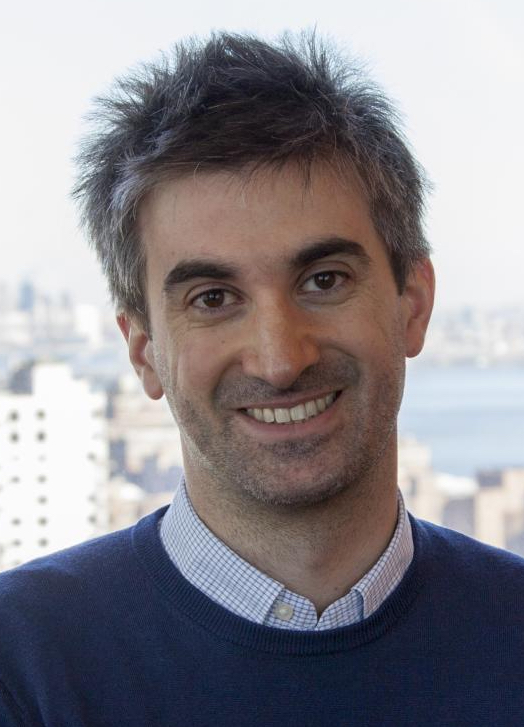}}]{Ludovic Righetti}
is an Associate Professor in the Electrical and Computer Engineering Department and in the Mechanical and Aerospace Engineering Department at the Tandon School of Engineering at New York University and a Senior Researcher at the Max-Planck Institute for Intelligent Systems in T\"ubingen, Germany. He holds an engineering diploma in Computer Science and a Doctorate in Science from the Ecole Polytechnique F\'ed\'erale de Lausanne (Switzerland). His research focuses on the planning and control of movements for autonomous robots, with a special emphasis on legged locomotion and manipulation.

\end{IEEEbiography}

\end{document}